\documentclass[10pt,twocolumn,letterpaper]{article}

\usepackage{cvpr}              

\usepackage[dvipsnames]{xcolor}

\definecolor{cvprblue}{rgb}{0.21,0.49,0.74}
\usepackage[pagebackref,breaklinks,colorlinks,citecolor=cvprblue]{hyperref}

\definecolor{MyDarkBlue}{rgb}{0,0.08,1}
\definecolor{MyDarkGreen}{rgb}{0.02,0.6,0.02}
\definecolor{CheckmarkGreen}{rgb}{0,1,0}
\definecolor{MyDarkRed}{rgb}{0.8,0.02,0.02}
\definecolor{MyDarkOrange}{rgb}
{0.40,0.2,0.02}
\definecolor{MyPurple}{RGB}{
111,0,255}
\definecolor{MyRed}{rgb}{1.0,0.0,0.0}
\definecolor{MyBlue}{rgb}{0.0,0.0,1.0}
\definecolor{MyGold}{rgb}{0.75,0.6,0.12}
\definecolor{MyDarkgray}{rgb}{0.66, 0.66, 0.66}

\newcommand{\oursabbv}[0]{Robo360}
\newcommand{\oursfull}[0]{A 3D Omnispective Multi-Material Robotic Manipulation Dataset}

\usepackage{pifont}  
\usepackage{boldline}
\usepackage{multirow}
\newcommand{\greencheck}[0]{\textcolor{MyDarkGreen}{\ding{51}}}
\newcommand{\redcross}[0]{\textcolor{MyDarkRed}{\ding{55}}}

\usepackage{float}
\usepackage{caption}

\def\paperID{2868} 
\def\confName{CVPR}
\def\confYear{2024}

\title{\oursabbv: \oursfull}
\author{
Litian Liang$^{1*}$
\ \ \ \ Liuyu Bian$^{2*}$
\ \ \ \ Caiwei Xiao$^1$
\ \ \ \ Jialin Zhang$^2$
\\
\ \ \ \ Linghao Chen$^1$
\ \ \ \ Isabella Liu$^1$
\ \ \ \ Fanbo Xiang$^1$
\ \ \ \ Zhiao Huang$^1$
\ \ \ \ Hao Su$^1$
\\
\\
$^1$UC San Diego
\ \ \ \
$^2$Tsinghua University
\\
*equal contribution
}

\newcommand{\myparagraph}[1]{\vspace{2mm}\noindent\textbf{#1}\quad}

\begin{document}
\twocolumn[{
    \renewcommand\twocolumn[1][]{#1}
    \maketitle
    \begin{center}
        \centering
        \captionsetup{type=figure}
        \includegraphics[width=\textwidth]{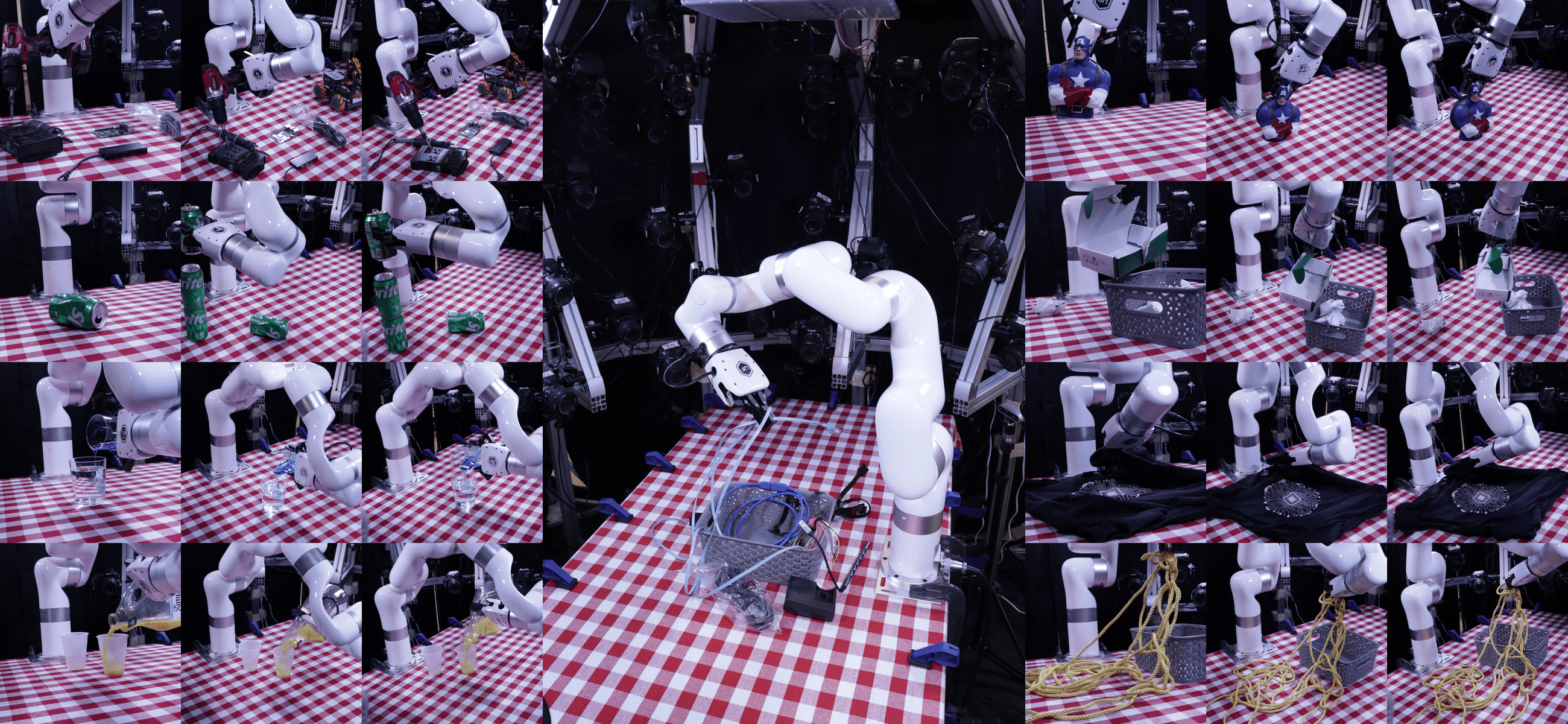}
        \caption{Example multi-view videos in \oursabbv \ dataset.}
        \label{tea}
    \end{center}
}]
\begin{abstract}
\vspace{-1.5em}
Building robots that can automate labor-intensive tasks has long been the core motivation behind the advancements in computer vision and the robotics community. 
Recent interest in leveraging 3D algorithms, particularly neural fields, has led to advancements in robot perception and physical understanding in manipulation scenarios. However, the real world's complexity poses significant challenges. To tackle these challenges, we present {\oursabbv}, a dataset that features robotic manipulation with a dense view coverage, which enables high-quality 3D neural representation learning, and a diverse set of objects with various physical and optical properties and facilitates research in various object manipulation and physical world modeling tasks. We confirm the effectiveness of our dataset using existing dynamic NeRF and evaluate its potential in learning multi-view policies. We hope that \oursabbv \ can open new research directions yet to be explored at the intersection of understanding the physical world in 3D and robot control.
\end{abstract}

\vspace{-0.5em}
\section{Introduction}

Mastering robotic manipulation in 3D is crucial for embodied agents. It involves manipulating objects within three-dimensional spaces to accomplish tasks. This field has seen growing interest recently. Advancements in 3D algorithms, particularly neural fields, have significantly enhanced robots' ability to perceive and act in 3D environments~\cite{ze2023visual,ling2023efficacy,li2023behavior,gu2023maniskill2}. Techniques like \cite{driess2022reinforcement,ze2023gnfactor,shim2023snerl} use neural fields for neural policy training, facilitating complex decision-making in 3D scenarios. Moreover,  \cite{li2021neural,driess2023learning,li2023pac,sanchez2020learning} employ these fields in world model developments~\cite{ha2018world,li2018learning,sanchez2020learning}.

Despite recent advancements, the real-world complexity poses substantial challenges to robotic systems. These challenges include the need for precise recognition of objects, materials, and deformations in 3D space, and the handling of complex scenarios involving specular and transparent objects. Additionally, real-world objects exhibit diverse physical properties, from rigid bodies to deformables and fluids. Consequently, creating comprehensive world models that support model-based planning and the development of generalizable policies is an intricate task.


To overcome these challenges, the development of \emph{a dataset of dynamic scenes featuring real-world robot manipulation} is crucial. This dataset would propel the development of algorithms and hardware in 3D robotic manipulation, serving as an exhaustive test set for evaluating progress. The dataset should include a diverse range of objects with different visual and physical attributes to enable robust testing across various scenarios. Moreover, it should provide abundant, and ideally, redundant 3D information to facilitate the comparison of robot vision systems with varying algorithmic and hardware configurations. 
With a one-time investment in collecting such information-rich data, which might be via advanced devices, benefits may be brought to meet varied user requirements. For example, the dataset can help design systems with limited views that are cheap and scalable at deployment time. As another example, the detailed information on deformable objects will aid in the training and evaluation of neural simulators~\cite{driess2023learning,pmlr-v164-li22a}.
Lastly, the dataset must incorporate action information so agents can understand how their actions influence the 3d environments.


Existing datasets often fail to meet these criteria. 3D datasets like~\cite{chang2017matterport3d} typically use limited viewpoint cameras and struggle with non-diffuse materials. Motion datasets like~\cite{lu2023diva,liu2022hoi4d} lack detailed action information. Large-scale robotic manipulation datasets~\cite{walke2023bridgedata,fang2023rh20t} still face viewpoint limitations, highlighting the need for a comprehensive approach combining robotics and 3D vision. This gap is also evident in fields like dynamic NeRF~\cite{fridovich2023k,yang2023deformable}, which often rely on limited real or simulated scenes for training.

In this paper, we introduce {\oursabbv}, showcased in Figure~\ref{tea}, as the first real-world omnispective robotic manipulation dataset specifically designed to foster research in 3D robotic manipulation. This dataset comprises over $2,000$ multi-view robotic manipulation trajectories generated through teleoperation~\cite{qin2023anyteleop}. To capture these trajectories, we established a multi-view system with $86$ cameras, each precisely calibrated and temporally aligned across different viewpoints. The extensive visual information in {\oursabbv} facilitates the creation of high-quality 3D neural fields, leveraging recent progress in 3D neural representation learning. The diversity of {\oursabbv} is another highlight; the dataset includes $100$ objects with over $20$ different materials, encompassing a broad spectrum of optical and physical properties. This diversity supports studies in a wide array of object manipulation and physical world modeling challenges. Comparative analysis of {\oursabbv} with previous real-world robotic datasets is presented in Tables~\ref{tab:multimodality}.


In preliminary experiments with {\oursabbv}, we explore 3D neural scene representation learning and multi-view policy learning to validate the dataset. We systematically evaluate dynamic neural radiance field methods on our dataset and confirm the dataset's effectiveness in supporting imitation learning for robots to develop and generalize multi-view manipulation skills. {\oursabbv} is expected to advance research in 3D scene representation learning, visual policy learning, and material modeling, enhancing the understanding of the physical world for robot control.

Our contributions are summarized as follows:
\begin{itemize}
    \item We curate a unique dataset featuring dynamic scenes of robot manipulation, encompassing dense view coverage, diverse objects with distinct mechanical and deformability properties, various optical materials, per-frame high-quality 3D neural representations, and precise robot action information.
    \item We detail the technical challenges encountered in building a multi-view robotic manipulation data collection system and outline a robust pipeline to address these challenges.
    \item We conduct evaluations of existing 3D dynamic NeRF algorithms in real-world robot manipulation scenarios.
    \item  A policy learning algorithm that adapts to multi-view inputs, showcasing its versatility across various views.
\end{itemize}


\section{Related Work}

\paragraph{3D Dataset Captured in Real World }

Capturing real-world objects and scenes is vital for computer vision research. Numerous works~\cite{silberman2012indoor,song2015sun,dai2017scannet,chang2017matterport3d,baruch2021arkitscenes} focus on collecting large-scale scenes for indoor 3D scene understanding, using RGB and depth cameras like Kinect and iPhone LiDAR scanners. Similarly,~\cite{singh2014bigbird,downs2022google} employ RGB-D camera rigs to capture individual objects, resulting in influential datasets like YCB~\cite{calli2015benchmarking}. However, existing depth sensors struggle to capture accurate depth for highly specular or transparent objects.
Unlike depth sensors, RGB cameras directly capture light and are unaffected by these issues. Some works~\cite{jensen2014large,toschi2023relight} use gantries for precise camera control during capture, but they have limited speed and are unsuitable for dynamic scenes. To capture dynamic scenes, complex view-dependent effects, and visual parameters, large arrays of RGB cameras and light stages have been employed. For instance, \cite{lombardi2019neural} uses a 34-camera array to capture intricate visual effects like fog, fluids, and fur, pioneering neural radiance capture. Meanwhile,~\cite{debevec2000acquiring,liu2023openillumination} concentrate on capturing objects under controlled lighting conditions for inverse rendering and material decomposition. Additionally,~\cite{peng2021neural,lu2023diva,joo2015panoptic} capture dynamic scene videos to advance dynamic scene representation in the field.

\paragraph{Dynamic NeRF}
The advancements in neural radiance fields (NeRF)~\cite{mildenhall2021nerf} have significantly improved the quality of novel view synthesis, enabling the generation of realistic and detailed images from 3D scenes. One notable extension of NeRF focuses on addressing dynamic scenes, where objects or the scene itself undergo temporal changes. This has led to the development of dynamic NeRF methods. One stream of the dynamic NeRF works directly extends the radiance field with the time dimension or a latent code \cite{du2021neural, gao2021dynamic, li2021neural, xian2021space}; some works utilize volume factorization to convert 4D volume into multiple planes or tensors~\cite{cao2023hexplane, fridovich2023k, shao2023tensor4d}, which provides a compact and efficient representation for dynamic scenes. Another direction of works~\cite{park2021nerfies, pumarola2021d, park2021hypernerf, tretschk2021non, fang2022fast}  focuses on constructing an additional deformation field that maps point coordinates from different time frames into a canonical space, where large motion and geometry changes can be captured and learned. Recently, with the emergence of 3D Gaussian Splatting~\cite{kerbl20233d}, which adopts a novel approach based on point cloud rendering, a series of works started using 3D Gaussians to model the dynamic scenes~\cite{yang2023deformable, wu20234d, luiten2023dynamic}.


\paragraph{3D Robot Learning}
There is growing interest in robotics policies based on 3D vision~\cite{ling2023efficacy,gu2023maniskill2,driess2022reinforcement,ze2023gnfactor}, and NeRFs derived from multi-view images have proven effective~\cite{ze2023gnfactor, driess2022reinforcement, pmlr-v164-li22a, wang2023d, driess2023learning}. These neural fields serve as 3D representations, trained concurrently with policy learning by reconstructing multi-view observations. Then, policies are trained to condition on past observations, using generative models to output action sequences~\cite{chi2023diffusion}. For example, imitation learning~\cite{ze2023gnfactor} learns policies from demonstrations, while reinforcement learning directly optimizes policies~\cite{driess2022reinforcement}.
Neural renderers also enhance world model learning~\cite{driess2023learning, seo2023multi, li2023pac}, allowing robots to plan using world models and forecast expected rewards. 

\paragraph{Dataset for Robotic Manipulation and Physics}
Recent advancements in imitation learning highlight the importance of large-scale robot datasets for policy learning. These datasets are gathered through various methods such as kinesthetic teaching~\cite{sharma2018multiple}, composing primitive actions~\cite{dasari2019robonet, brohan2022rt}, or teleoperation~\cite{walke2023bridgedata, fang2023rh20t}. While some datasets include multiple views~\cite{dasari2019robonet, walke2023bridgedata, fang2023rh20t}, their limited viewpoints hinder comprehensive 3D scene understanding.
In contrast, obtaining multi-view and 3D data is more straightforward in simulated environments~\cite{mandlekar2018roboturk, mandlekar2021matters, mees2022calvin, james2020rlbench, li2023behavior, gu2023maniskill2, yu2020meta, mandi2022towards}, which facilitate testing robot learning and advancing physical world modeling, intuitive physics, and reasoning research~\cite{li2018learning, bakhtin2019phyre, bear2021physion}. Despite some progress in soft body simulation~\cite{huang2021plasticinelab, xian2023fluidlab, lin2021softgym}, if they can match the simulator with real-world dynamics remains challenging. A dataset encompassing various physical dynamics that can serve as a benchmark for evaluating these simulators becomes essential.

\begin{figure*}[ht]
    \includegraphics[width=\textwidth
]{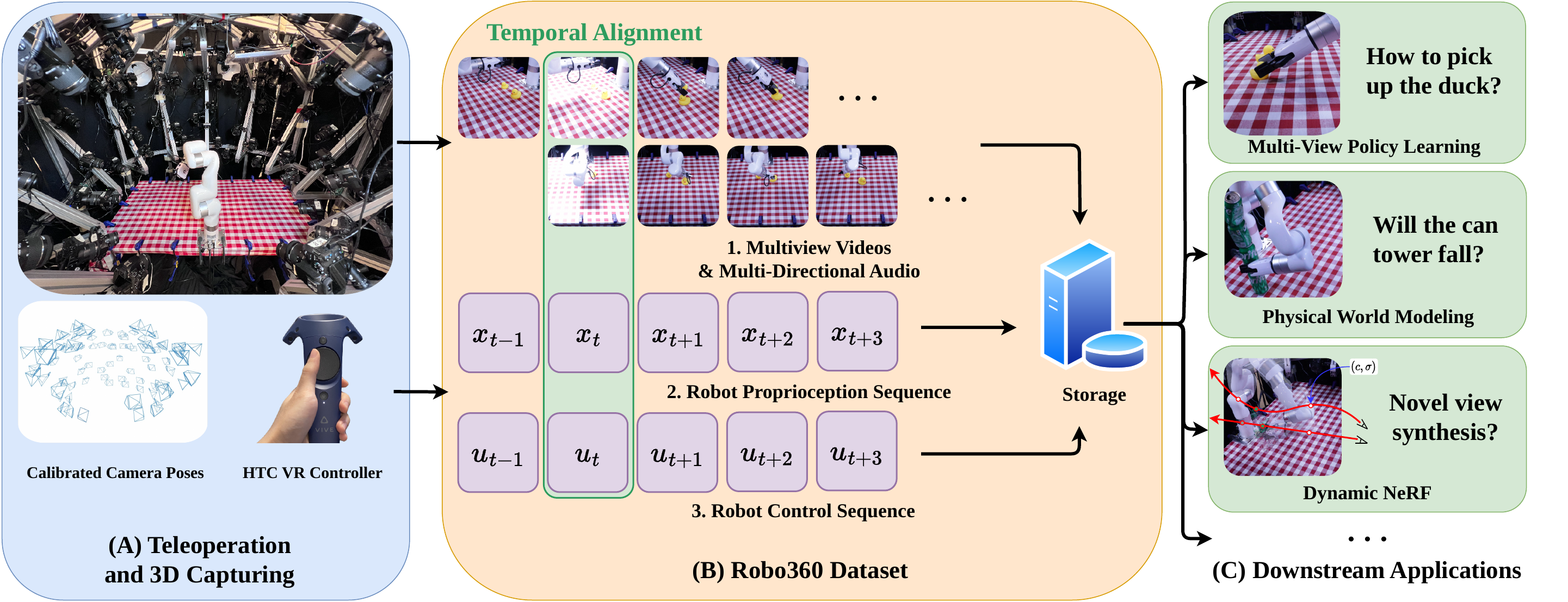}
    \caption{An illustration of the comprehensive pipeline encompassing the stages of data collection, postprocessing, storage, and the subsequent downstream applications.}
    \label{fig:pipeline}
\end{figure*}

\begin{center}
\begin{table*}[ht]
\centering
\renewcommand{\arraystretch}{1.1}
\resizebox{\textwidth}{!}{
\begin{tabular}{l|ccccccc|ccc}
\hlineB{2}
\textbf{Name} & \textbf{Robot} & \textbf{Control} &\textbf{\# Views} & \textbf{Resolution} & \textbf{Camera Calib.} & \textbf{Audio} & \textbf{Depth}  & \textbf{Cables} & \textbf{Cloth} & \textbf{Fluid}\\
\hlineB{2}
Deep3DMV \citep{lin2021deep}
& \redcross & \redcross & 10 & \textbf{1080p} & \greencheck & \redcross & \redcross & \redcross & \greencheck & \redcross \\
DiVA360 \citep{lu2023diva}
& \redcross & \redcross  & 53 & 720p & \greencheck & \greencheck & \redcross & \redcross & \redcross & \redcross\\
RoboSet \citep{bharadhwaj2023roboagent}
& \greencheck & 5 Hz & 4 & 480p & \redcross & \redcross & \greencheck & \redcross & \redcross & \redcross\\
BridgeDataV2 \citep{walke2023bridgedata}
& \greencheck & 5 Hz & 4 & 480p & \redcross & \redcross & \greencheck & \redcross & \redcross & \redcross\\
RH20T \citep{fang2023rh20t} 
& \greencheck & 10 Hz & 12 & 720p & \greencheck & \greencheck & \greencheck & \greencheck & \greencheck & \greencheck \\
\hline
\oursabbv \ \textbf{(Ours)}
& \greencheck & \textbf
{30 Hz} & \textbf{86} & \textbf{1080p} & \greencheck & \greencheck & \greencheck & \greencheck & \greencheck & \greencheck \\
\hlineB{2}
\end{tabular}
}
\caption{Comparing our dataset with existing real-world multi-view video datasets.}
\label{tab:multimodality}
\end{table*}
\end{center}


\section{\oursabbv \ Dataset}



%
%


\oursabbv \ is the first real-world multi-view dataset that simultaneously supports high-quality dynamic 3D scene representation and learning-based robot control. {\oursabbv} aims to support research in 3D robotic manipulation, especially focusing on understanding the low-level dynamics in the manipulation process. To achieve this, we include multi-modality data in the \oursabbv \ dataset.
Specifically, \oursabbv \ contains over 2000 demonstration trajectories of more than 100 different objects with diverse material variations. The multi-view trajectory RGB video and multi-directional audio are captured by $86$ DSLR cameras, along with data with different modalities, including robot proprioception and robot control signal. This section highlights several notable features of {\oursabbv}.

\myparagraph{RGB Video Sequences}
The robot execution trajectory video is captured with 86 DSLR cameras distributed across a half-dome. We calibrated the intrinsic and extrinsic parameters of all cameras.
The visualization of the dome and calibrated cameras are in Figure \ref{fig:pipeline} (A).
All videos are captured at 30 FPS with 1080p resolution.

\myparagraph{Multi-Material}
\begin{figure*}[t]
    \centering
    \includegraphics[width=\textwidth]{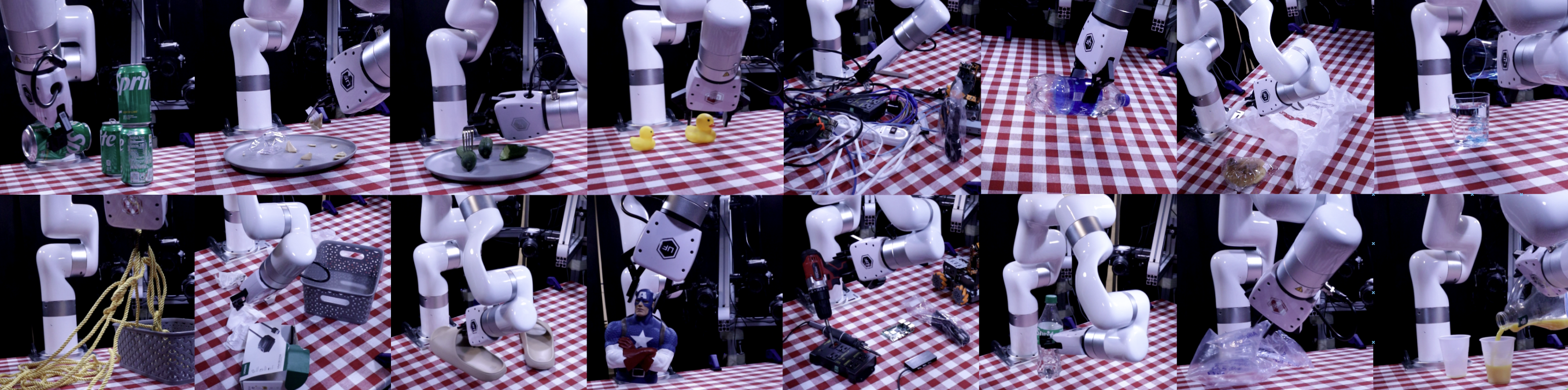}
    \caption{{\oursabbv} captures real-world robot-object interactions with complex visual and material variations. }
    \label{fig:material_diversity}
\end{figure*}
Our dataset, \oursabbv, captures diverse interactions between robots and objects with a wide range of visual and physical characteristics, as depicted in Figure \ref{fig:material_diversity}. It includes over 200 trajectories for each object of rigid material, including various plastic objects, wood, and metals. These objects typically undergo rigid transformations; however, some of them, such as biscuits, can experience fractures during interaction. The dataset also encompasses soft materials, including rubber, cloth, paper, and malleable metals like copper and iron, with over 200 trajectories for each. Many of the soft objects in our dataset are composites, such as cardboard boxes, napkins, electronic cables, plastic bags, and ropes, combining these materials in varied configurations.
Additionally, certain soft objects, such as fruits, snacks, and miscellaneous items, combine a distinctive set of materials that distinguishes them from other objects. Regarding liquids, the dataset encompasses a variety of types characterized by different colors, levels of transparency, densities, viscosities, and surface tensions. This includes substances such as water, fruit juice, dish soap, soda, and coffee for more than 30 trajectories in total.

\myparagraph{Robot Proprioception and Control}
{\oursabbv} captures high-frequency robot proprioceptions (joint position, joint velocity, joint acceleration, and joint torque) at 30 Hz, and teleoperation signals (target joint position, target gripper position) at 30 Hz. We also record the analytically solved end-effector pose and gripper tip position for the convenience of downstream applications.

\myparagraph{Multi-Modality}
To support research in understanding the physical world, \oursabbv \ consists of a rich set of different modalities, including
86-directional audio collected by DSLR cameras at 48000 Hz and depth map collected by 3 RealSense depth cameras.


\section{RoboStage for Dataset Creation}


We develop a multi-view system with $86$ cameras to capture rich multi-modal data of the robot manipulation trajectories to support research in 3D robotic manipulation. Inspired by light stages in graphics~\cite{lombardi2019neural,liu2023openillumination}, we built our hardware and software system, which we call RoboStage. RoboStage can capture high-resolution multi-view videos and audio and synchronize them with the control signal of the robot arm. By leveraging a VR controller, we built a teleoperation system inside the RoboStage to collect robotic manipulation trajectories.  Figure~\ref{fig:pipeline} shows the system and illustrates the dataset creation process, which enables collecting data that supports multiple downstream applications.

\subsection{RoboStage Setup}
RoboStage, as shown in Figure~\ref{fig:pipeline} (A), consists of 86 Canon DSLR 250D cameras (DSLR), and 3 RealSense depth cameras.  The DSLR cameras are mounted on the upper half dome supported by a $2m \times 2m \times 2m$ metal frame, allowing RoboStage to capture videos from different directions. The system's specifics can be found in the supplementary.



\subsection{Dynamic 3D Scene Capture}
We carefully calibrate cameras to ensure precise parameters and align videos to ensure synchronized timesteps.

\myparagraph{Camera Calibration}
Accurate camera parameters play a pivotal role in a multi-view dataset, particularly for tasks like 3D scene learning and novel-view synthesis.
In our capturing system, where the positions and orientations of all cameras are fixed, we opt to perform calibration before the capturing process, mitigating the need for per-scene re-calibration.
Specifically, we create a static scene with rich texture and low specularity. Subsequently, we employ COLMAP~\cite{schonberger2016pixelwise} to estimate camera parameters. To obtain the metric scales and coordinate uniformity among camera poses, we utilize an ArUco board~\cite{garrido2014automatic} to estimate the poses of a subset of the cameras and align the results from COLMAP and ArUco. We illustrated the calibrated camera poses in Figure~\ref{fig:pipeline} (A).

\myparagraph{Video Temporal Alignment}
To support applications in the vision and robotics community, such as dynamic NeRF and imitation learning, we align the multi-view video, robot proprioception, and robot control sequences in the time dimension.
To this end, we implemented a common event, i.e., a floodlight, that is easily detectable at the beginning of the video, serving as a keyframe to synchronize all videos.
At the start of each trajectory, we activate a 30000-lumen floodlight and maintain its illumination for approximately 3 seconds. Subsequently, we calculate the average pixel value difference between each pair of consecutive frames. The frame exhibiting the maximum difference is identified as the keyframe.
In addition to aligning the cameras, we align control signals, robot proprioception, and video frames using the RealSense cameras as a proxy. In detail, during each teleoperation step, we concurrently record the control signals and robot proprioception while capturing images using the RealSense cameras. This simultaneous recording aligns the control signals, robot proprioception, and RealSense videos automatically. Then, by applying the method described earlier to align the RealSense videos with the DSLR videos, we can achieve synchronization for the control signals and robot proprioception as well.
We verified our alignment approach by comparing it with an iPad displaying a QR code video at 30 FPS in front of each camera and observing a maximum discrepancy of 1/30 seconds, while our method offers the advantage of greater automation as it is much cheaper and easier to activate.


\subsection{Robot Arm}
An xArm6 robot arm, equipped with 6 revolute joints and a parallel gripper, is mounted at the center of the capturing system. During data collection, the robot is teleoperated by a human operator, while during deployment, it can be controlled by a policy neural network. In this dataset, we use servo joint position as the low-level control signal for both teleoperation and policy deployment.

\myparagraph{Robot Teleoperation} The human teleoperator controls the robot using an HTC Vive VR gaming controller. During teleoperation, we use steam-openvr~\cite{Welcome_to_Steam_2020} to read the $3\times4$ target pose matrix and the trigger button status of the controller in real-time. Using inverse kinematics, we compute the closest target joint position to the current joint position that matches the end-effector pose to the target pose. We then compute the target gripper position by scaling the continuous-valued trigger position. Then, the robot arm and gripper are driven to the target joint position by making an API call to the xArm control box. The computed 6D target joint position and 1D target gripper position during teleoperation are recorded in the dataset, together with the real-time proprioception state information of the robot arm (joint position, joint velocity, joint acceleration, and joint torque). 

\myparagraph{Robot Safety} We provide safety measures to prevent the robot from causing damage to other devices, such as cameras, desks, and single-board computers. To ensure the integrity of extended data-capturing and policy evaluation sessions, we have implemented a straightforward bounding box collision check during both teleoperation and policy deployment.
Before directing the robot to a target joint position, we assess the positions of a predefined set of points on the surface of the robot links. These positions are then enclosed within a pre-defined bounding box. Subsequently, we calculate safe joint positions using inverse kinematics and execute these safe joint positions. This precautionary process mitigates the risk of collisions and safeguards against potential damage to surrounding devices.

\myparagraph{Policy Deployment}
The trained visuomotor neural network policy predicts the delta joint position and target gripper position to control the robot, given real-time robot proprioception and image observation from a subset (4 in our experiments) of 86 DSLRs.

We find that our data collection system is quite efficient and easy to learn. It took, on average, $2$ minutes to collect a $30$-seconds manipulation trajectory, including scene setup, temporal alignment, object manipulation, and data transfer. Moreover, our system enables a higher control frequency than other datasets in Table~\ref{tab:multimodality}, supporting more complicated manipulation skills. We leave more details in the supplementary.

\begin{center}
\begin{table}[t]
\centering
\renewcommand{\arraystretch}{1.2}
\resizebox{\linewidth}{!}{
\begin{tabular}{|c|ccc|cc|}
\hlineB{2}
Method & D-NeRF & K-Planes & De. 3DGS & TensoRF &  3DGS  \\ 
\hline
Per-frame & \redcross & \redcross & \redcross & \greencheck & \greencheck \\
\hlineB{2}
\textbf{Bottle Flip}     & 10.96   & 16.94    & \textbf{18.32}    & 19.42      & \textbf{24.84}    \\ 
\textbf{Bread Fall}      & 10.75   & 17.32    & \textbf{19.05}    & 20.36      & \textbf{24.90}    \\ 
\textbf{Cucumber Skewer} & 11.02   & 17.21    & \textbf{18.87}    & 20.05      & \textbf{25.36}    \\ 
\textbf{Dish Soap}       & 10.82   & 16.91    & \textbf{19.05}    & 21.05      & \textbf{24.90}    \\ 
\textbf{Tshirt Fold}     & 11.27   & 16.89    & \textbf{18.52}    & 20.36      & \textbf{24.07}    \\ 
\textbf{Cookie Crushed}  & 11.35   & 17.10    & \textbf{18.75}    & 20.89      & \textbf{25.48}    \\ 
\textbf{Rope Pick-and-Place}    & 10.39   & 16.85    & \textbf{18.65}    & 20.47      & \textbf{24.70}    \\ 
\textbf{Air-bag Squeeze} & 10.46 & 17.31    & \textbf{17.94}    & 20.55      & \textbf{24.06}    \\
\textbf{power strip}     & 10.40   & 16.74    & \textbf{18.09}    & 20.08      & \textbf{23.90}    \\ 
\textbf{Juice Pour}      & 11.13   & 17.08    & \textbf{18.59}    & 20.60      & \textbf{24.20}    \\ 
\textbf{Bread Put-in-bag}       & 9.94    & 17.07    & \textbf{17.78}    & 19.79      & \textbf{23.94}    \\ 
\textbf{Sand Pour}       & 10.88   & 17.26    & \textbf{18.97}    & 20.65      & \textbf{25.78}    \\ 
\textbf{Sprite Pour}     & 10.21   & 17.14    & \textbf{18.91}    & 19.79      & \textbf{25.64}    \\ 
\textbf{Sprite Stack}    & 10.09   & 17.12    & \textbf{18.83}    & 20.46      & \textbf{24.74}    \\ 
\textbf{Tshirt Unfold}   & 10.91   & 17.15    & \textbf{18.03}    & 20.01      & \textbf{24.18}    \\
\hlineB{2}
\end{tabular}
}
\caption{Dynamic NeRF Evaluation: Average PSNR across a 5-second video.}
\label{tab:dnerf}
\end{table}
\end{center}

\begin{table}[h]
    \centering
    
    \resizebox{0.9\linewidth}{!}{
    \begin{tabular}{|c|c|cccc|}

        \hline
        \textbf{Task} & \textbf{Test} & \textbf{BC(F)} & \textbf{BC(A)} & \textbf{DP(F)} & \textbf{DP(A)} \\
        \hline
        \multirow{4}{*}{\textbf{Towel}} 
        & Train  & 0.60 & 0.87 & \textbf{1.00} & \textbf{1.00} \\
        & 1-Test & 0.26 & \textbf{0.80} & 0.40 & \textbf{0.80} \\
        & 2-Test & 0.00 & 0.73 & 0.00 & \textbf{0.80} \\
        & 3-Test & 0.00 & 0.67 & 0.00 & \textbf{0.73} \\
        \hline
        \multirow{4}{*}{\textbf{Slippers}} 
        & Train  & 0.20 & 0.00 & \textbf{0.53} & \textbf{0.53} \\
        & 1-Test & 0.07 & 0.00 & 0.00 & \textbf{0.60} \\
        & 2-Test & 0.00 & 0.00 & 0.00 & \textbf{0.53} \\
        & 3-Test & 0.00 & 0.00 & 0.00 & 0.00 \\
        \hline
        \multirow{4}{*}{\textbf{Bottle}} 
        & Train  & 0.27 & 0.80 & \textbf{1.00} & 0.87 \\
        & 1-Test & 0.00 & 0.33 & 0.20 & \textbf{0.87} \\
        & 2-Test & 0.00 & 0.33 & 0.00 & \textbf{0.80} \\
        & 3-Test & 0.00 & 0.13 & 0.00 & \textbf{0.80} \\
        \hline
        \multirow{4}{*}{\textbf{Cable}} 
        & Train  & 0.47 & 0.60 & \textbf{0.87} & 0.67 \\
        & 1-Test & 0.00 & \textbf{0.60} & 0.00 & \textbf{0.60} \\
        & 2-Test & 0.00 & 0.27 & 0.00 & \textbf{0.47} \\
        & 3-Test & 0.00 & 0.00 & 0.00 & 0.00 \\
        \hline
        \multirow{4}{*}{\textbf{Rope}} 
        & Train  & 0.67 & \textbf{1.00} & \textbf{1.00} & \textbf{1.00} \\
        & 1-Test & 0.00 & 0.60 & 0.47 & \textbf{0.80} \\
        & 2-Test & 0.00 & 0.00 & 0.00 & \textbf{0.67} \\
        & 3-Test & 0.00 & 0.00 & 0.00 & 0.00 \\
        \hline
    \end{tabular}
    }
    \caption{Imitation learning for robot control baseline methods task success rate. We report the average success rate over 15 trials. BC and DP denote Behavior Cloning and Diffusion Policy, respectively. (F) denotes the method is trained on a fixed 4 views. (A) denotes the method is trained on randomly sampled 4 views. We test the methods in different 4-view combinations, where ``Train'' means the 4 training views are used as input during deployment. ``N-Test'' means we use N novel views and 4-N training views as input during deployment.}
    \label{tab:imitation_learning_results}
\end{table}
\vspace{-3em}
\section{Baseline Experiments}

In this section, we verify our dataset in two tasks: novel-view synthesis of dynamic scenes and policy imitation learning. 
We split our dataset into two categories for these two tasks.
For novel-view synthesis of dynamic scenes, we use videos focusing on collecting a diverse set of objects with visual and material variation, while for policy learning, we use videos focusing on a small set of objects with each 50-200 trajectories.
We found that from \oursabbv \ we can obtain high-quality neural radiance fields, and the collected demonstrations can efficiently drive robot control policy learning, proving the effectiveness of our dataset. We systematically evaluated the existing 3D dynamic NeRF methods in our dataset and found that existing dynamic scene representation learning methods are insufficient to model objects with fast motion, resulting in a performance gap compared to static scenes.

\subsection{Dynamic NeRF}

We selected 15 different scenes to compare various dynamic NeRF methods. 
These scenes cover a wide range of scenarios involving the robot's interactions with objects that possess diverse materials, both in terms of their visual and physical properties, 
We crop the manipulation process into a 5-second 30FPS video, yielding a total of $150$ frames per scene. For evaluating the models' ability in novel-view synthesis, we hold $5$ of $86$ views as the test set.

\begin{figure*}[htp]
    \includegraphics[width=0.8\textwidth]{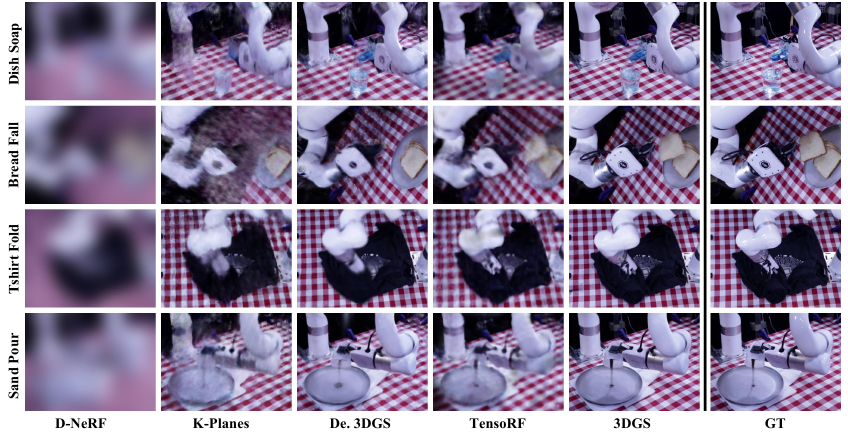}
    \centering
    \caption{Qualitative results of dynamic NeRF methods. All the dynamic NeRF methods fail to accurately model objects with fast motion, such as falling bread and pouring sand, leading to a performance gap compared to static scenes.}
    \label{fig:dnerf}
    \vspace{-0.5cm}
\end{figure*}

\paragraph{Baselines}

We validate five different dynamic NeRF methods: D-NeRF\cite{pumarola2021d}, K-Planes\cite{fridovich2023k}, Deformable 3DGS\cite{yang2023deformable}, per-frame TensoRF\cite{chen2022tensorf}, and per-frame 3DGS\cite{kerbl20233d}. We report the PSNR results of these methods on novel views in Table~\ref{tab:dnerf} and provide qualitative results in Figure \ref{fig:dnerf}.

D-NeRF employs a Multilayer Perceptron (MLP) to model the dynamic scene by directly extending the radiance field with the time dimension as an input. Thus, its performance is constrained by the network capacity, particularly in scenarios with a large number of frames.
Despite being trained for 10 hours per scene on our dataset, D-NeRF can only capture a fuzzy representation of scene composition, lacking detailed information.

K-Planes utilizes six planes to decompose spatial-temporal volumes. In our dataset, K-Planes exhibits strong performance in areas visible from most views but struggles in regions visible from only a small subset of cameras. Since our cameras are predominantly oriented toward the center of the light stage, where most motion occurs, K-Planes accurately depicts the approximate motion of the object in these regions. However, the novel-view renderings lack temporal smoothness due to the limited representation ability.

Deformable 3DGS adds a deformation field to Gaussian Splatting to enable dynamic scene novel-view synthesis.
While it represents the state-of-the-art method in dynamic NeRF, it still exhibits several limitations when applied to our dataset.
First, the 5-second video contains too much information and is challenging for the capacity of Deformable 3DGS.
Second, we also observe that the GS-based methods fail and produce a worse result on the specular surfaces.

In addition to the dynamic NeRF methods, we also validate TensoRF and Gaussian Splatting for novel-view synthesis methods in static scenes by training and testing one model frame by frame. 
We restrict the model training on each frame for a maximum duration of 5 minutes to avoid exceeding the computation budgets.
We notice TensoRF focuses on modeling the whole scene and achieves a relatively high PSNR but lacks details on the manipulated object. 
For Gaussian Splatting, we initialized the Gaussian using COLMAP and trained the model for $10000$ iterations. We noticed an increase in the degree of spherical harmonic function boosted the performance on the specular region of the robot surface. Finally, per-frame Gaussian Splatting delivers the highest PSNR in our dataset in all scenes. 

We also conducted an ablation study to compare their performances under different total lengths of frames. Please refer to the supplementary material for more results.


\subsection{Learning-Based Robot Control}
\begin{figure*}[h]
    \centering
    \begin{subfigure}[b]{0.19\textwidth}
        \includegraphics[width=\textwidth]{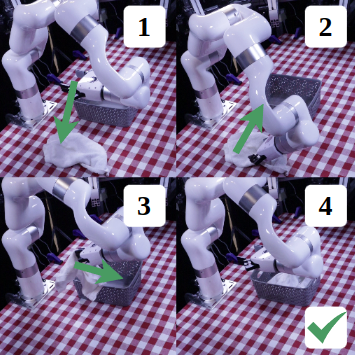}
        \caption{Towel}
    \end{subfigure}
    \hfill
    \begin{subfigure}[b]{0.19\textwidth}
        \includegraphics[width=\textwidth]{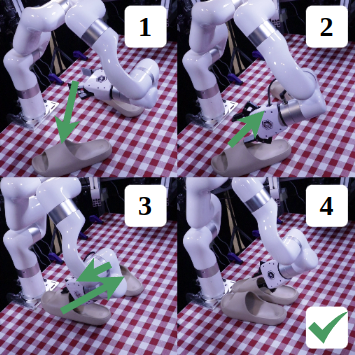}
        \caption{Slippers}
    \end{subfigure}
    \hfill
    \begin{subfigure}[b]{0.19\textwidth}
        \includegraphics[width=\textwidth]{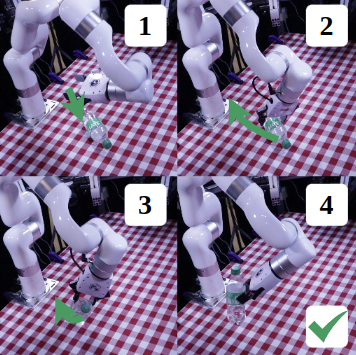}
        \caption{Bottle}
    \end{subfigure}
    \hfill
    \begin{subfigure}[b]{0.19\textwidth}
        \includegraphics[width=\textwidth]{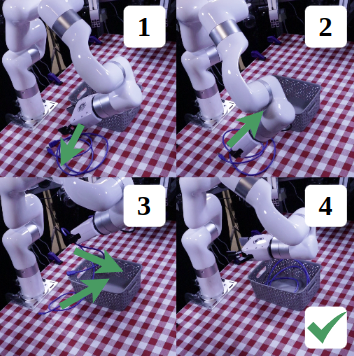}
        \caption{Cable}
    \end{subfigure}
    \hfill
    \begin{subfigure}[b]{0.19\textwidth}
        \includegraphics[width=\textwidth]{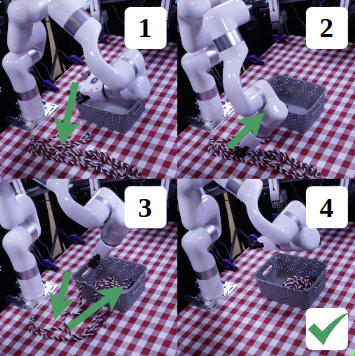}
        \caption{Rope}
    \end{subfigure}
    \caption{
    Visualization of 5 example demos used in training networks via imitation learning. For each demo, we show 4 keyframes, where  \textcolor{MyDarkGreen}{green arrows} show the desired end-effector motion and \textcolor{MyDarkGreen}{green checkmarks} denote the task is completed at the current time. Zoom in for details.
    }
    \label{fig:task_descriptions}
\end{figure*}
We now verify if our dataset can be used to learn robot control policies with imitation learning and if our dataset can help learned policies generalize to novel views. We selected five representative soft body manipulation tasks involving distinct materials: cloth, rubber, soft metal, plastic, and liquid. The specifics of these tasks are illustrated in Figure \ref{fig:task_descriptions}. Our analysis used a targeted subset of our dataset, comprising 30-200 demonstrations for each object. Specifically, we gathered 80 demos for Towel, 30 for Slippers, 80 for Bottle, 150 for Cable, and 155 for Rope to train the policy networks. We design the policy network to receive and process two sequential observations of the task, each from four different angles, presented in $128\times128$ RGB images. It outputs a delta joint position and a gripper position to control the robot.
\vspace{-1em}
\paragraph{Baseline Methods and Evaluation Metric}
We experimented with 2 imitation learning methods: Behavior Cloning (BC)~\cite{torabi2018behavioral} and Diffusion Policy (DP) \citep{chi2023diffusion}. BC learns the control policy by minimizing squared L2 loss between predicted action and ground truth action. Diffusion Policy learns a denoising network by predicting the noise from the action diffusion process. For each method, we train 2 different variants. The fixed-view network is denoted with suffix \textbf{(F)} and the arbitrary view network is denoted with suffix \textbf{(A)}. We verify the correctness of our dataset by training and evaluating the fixed-view network and then understand how our dataset helps with learning a policy that can generalize to novel view directions with the arbitrary view network. The fixed-view network computes the multi-view image feature with a convolutional neural network (CNN) with input channel 24 (4 RGB views, 2 consecutive frames stacked). The arbitrary view network computes the multi-view image feature with a CNN image tokenizer with input channel 6 (1 RGB image, 2 consecutive frames stacked) and a transformer encoder~\cite{vaswani2017attention} that is used to process these tokens. For BC, the image features are processed by an MLP to output the action at the current timestep. For DP, the image features are processed by a transformer decoder to compute the denoising step of a sequence of actions. 
\vspace{-1.5em}
\paragraph{Experiment Results}
We evaluate the performance of baseline methods in Table \ref{tab:imitation_learning_results}. We measure the success rate of each method-network pair in 4 view combinations (Train, 1-Test, 2-Test, and 3-Test) as mentioned in the table. We first observe that, given in-distribution training views, both versions of DP outperform BC in most cases. In this setting, the fixed-view version performs similarly or better than the arbitrary-view version. However, in novel view control setting, for both BC and DP, once even 1 camera position is perturbed to a novel viewing direction, we observe a severe performance drop for the fixed-view policies, whereas the policy trained on arbitrary view data only drops slightly or maintains similar performance. This is because when a view is perturbed, the input is out-of-distribution (OOD) for fixed-view policies, while the arbitrary view policy can interpolate the novel view input between the training view data that is close to the perturbed view direction to generalize to these unseen view directions and maintain performance. In the novel view control setting, we also observe that arbitrary view DP is more robust to view perturbation than BC. This is likely because the diffusion denoising process itself can help with combating OOD. However, this is not sufficient for a fixed-view DP to generalize to the novel view directions, since in this case, the novel view image observation that the diffusion process is conditioned on is itself OOD to the network. For arbitrary view policies, the performance continues to drop in most cases when more views are perturbed. We want to emphasize that building robust view independent policies is important for many applications and remains an open problem.

The amount of data available to train a network is crucial to the performance of a policy in all tasks. We observe all methods perform the worst in the task Slippers, comparing to their performance in other tasks. The performance of arbitrary view BC is impacted the most as it completely fails in all trials of all view combinations. This is very likely since we only provided 30 demonstrations for this task, compared to other tasks. This shows the importance of collecting more diverse data for learning-based robot controllers. 

One important contributing factor to the performance of all trained policies is that the inference frequency needs to match the training frequency to achieve reasonable performance. We conjecture that it is because a significant delay would change the distribution of the observation, leading to a performance drop. We find that the inference time of BC is much shorter than DP, which does not require multi-step control sequence prediction to match the training data distribution. On the other hand, DP relies on multi-step prediction since the denoising step is relatively more compute-intensive. We find slightly reducing the number of transformer decoder layers of the denoising network benefits performance by providing a higher inference frequency.




\section{Conclusion and Limitation}


We introduce {\oursabbv}, a 3D omnispective multi-material robotic manipulation dataset. It comprises a diverse range of robot manipulation trajectories collected by teleoperation, encompassing various physical and optical properties, as well as manipulation skills. To facilitate data collection, we have constructed RoboStage, a multi-view system equipped with 86 cameras, enabling the collection of multi-modal observations with rich 3d information. 
However, certain limitations are inherent in our approach. Firstly, our heavy reliance on visual data limits our dataset's ability to discern the internal structure of objects, and it is susceptible to significant occlusion. Nonetheless, we believe this limitation aligns with the current state of vision-based robot control and is not necessarily a drawback. Additionally, it is worth noting that the setup of RoboStage entails non-negligible costs, which may impede its scalability. Although we anticipate expanding our data collection efforts in the future, an exciting avenue for further exploration lies in developing techniques aimed at learning 3D scenes more effectively using fewer views from our dataset.

{
    \small
    \bibliographystyle{ieeenat_fullname}
    \bibliography{main}

\begin{thebibliography}{75}
\providecommand{\natexlab}[1]{#1}
\providecommand{\url}[1]{\texttt{#1}}
\expandafter\ifx\csname urlstyle\endcsname\relax
  \providecommand{\doi}[1]{doi: #1}\else
  \providecommand{\doi}{doi: \begingroup \urlstyle{rm}\Url}\fi

\bibitem[Wel(2020)]{Welcome_to_Steam_2020}
Steam openvr, \href{https://store.steampowered.com/news/app/250820/view/2898585530113913716}{Welcome to Steam}, 2020.

\bibitem[Bakhtin et~al.(2019)Bakhtin, van~der Maaten, Johnson, Gustafson, and Girshick]{bakhtin2019phyre}
Anton Bakhtin, Laurens van~der Maaten, Justin Johnson, Laura Gustafson, and Ross Girshick.
\newblock Phyre: A new benchmark for physical reasoning.
\newblock \emph{Advances in Neural Information Processing Systems}, 32, 2019.

\bibitem[Baruch et~al.(2021)Baruch, Chen, Dehghan, Dimry, Feigin, Fu, Gebauer, Joffe, Kurz, Schwartz, and Shulman]{baruch2021arkitscenes}
Gilad Baruch, Zhuoyuan Chen, Afshin Dehghan, Tal Dimry, Yuri Feigin, Peter Fu, Thomas Gebauer, Brandon Joffe, Daniel Kurz, Arik Schwartz, and Elad Shulman.
\newblock Arkitscenes: A diverse real-world dataset for 3d indoor scene understanding using mobile rgb-d data.
\newblock 2021.

\bibitem[Bear et~al.(2021)Bear, Wang, Mrowca, Binder, Tung, Pramod, Holdaway, Tao, Smith, Sun, et~al.]{bear2021physion}
Daniel~M Bear, Elias Wang, Damian Mrowca, Felix~J Binder, Hsiao-Yu~Fish Tung, RT Pramod, Cameron Holdaway, Sirui Tao, Kevin Smith, Fan-Yun Sun, et~al.
\newblock Physion: Evaluating physical prediction from vision in humans and machines.
\newblock \emph{arXiv preprint arXiv:2106.08261}, 2021.

\bibitem[Bharadhwaj et~al.(2023)Bharadhwaj, Vakil, Sharma, Gupta, Tulsiani, and Kumar]{bharadhwaj2023roboagent}
Homanga Bharadhwaj, Jay Vakil, Mohit Sharma, Abhinav Gupta, Shubham Tulsiani, and Vikash Kumar.
\newblock Roboagent: Generalization and efficiency in robot manipulation via semantic augmentations and action chunking, 2023.

\bibitem[Brohan et~al.(2022)Brohan, Brown, Carbajal, Chebotar, Dabis, Finn, Gopalakrishnan, Hausman, Herzog, Hsu, et~al.]{brohan2022rt}
Anthony Brohan, Noah Brown, Justice Carbajal, Yevgen Chebotar, Joseph Dabis, Chelsea Finn, Keerthana Gopalakrishnan, Karol Hausman, Alex Herzog, Jasmine Hsu, et~al.
\newblock Rt-1: Robotics transformer for real-world control at scale.
\newblock \emph{arXiv preprint arXiv:2212.06817}, 2022.

\bibitem[Calli et~al.(2015)Calli, Walsman, Singh, Srinivasa, Abbeel, and Dollar]{calli2015benchmarking}
Berk Calli, Aaron Walsman, Arjun Singh, Siddhartha Srinivasa, Pieter Abbeel, and Aaron~M. Dollar.
\newblock Benchmarking in manipulation research: The ycb object and model set and benchmarking protocols.
\newblock \emph{IEEE Robotics \& Automation Magazine, 22 (2015) 36 - 52}, 22\penalty0 (3):\penalty0 36--52, 2015.

\bibitem[Cao and Johnson(2023)]{cao2023hexplane}
Ang Cao and Justin Johnson.
\newblock Hexplane: A fast representation for dynamic scenes.
\newblock In \emph{Proceedings of the IEEE/CVF Conference on Computer Vision and Pattern Recognition}, pages 130--141, 2023.

\bibitem[Chang et~al.(2017)Chang, Dai, Funkhouser, Halber, Nießner, Savva, Song, Zeng, and Zhang]{chang2017matterport3d}
Angel Chang, Angela Dai, Thomas Funkhouser, Maciej Halber, Matthias Nießner, Manolis Savva, Shuran Song, Andy Zeng, and Yinda Zhang.
\newblock Matterport3d: Learning from rgb-d data in indoor environments.
\newblock 2017.

\bibitem[Chen et~al.(2022)Chen, Xu, Geiger, Yu, and Su]{chen2022tensorf}
Anpei Chen, Zexiang Xu, Andreas Geiger, Jingyi Yu, and Hao Su.
\newblock Tensorf: Tensorial radiance fields.
\newblock In \emph{European Conference on Computer Vision}, pages 333--350. Springer, 2022.

\bibitem[Chi et~al.(2023)Chi, Feng, Du, Xu, Cousineau, Burchfiel, and Song]{chi2023diffusion}
Cheng Chi, Siyuan Feng, Yilun Du, Zhenjia Xu, Eric Cousineau, Benjamin Burchfiel, and Shuran Song.
\newblock Diffusion policy: Visuomotor policy learning via action diffusion.
\newblock \emph{arXiv preprint arXiv:2303.04137}, 2023.

\bibitem[Dai et~al.(2017)Dai, Chang, Savva, Halber, Funkhouser, and Niessner]{dai2017scannet}
Angela Dai, Angel~X. Chang, Manolis Savva, Maciej Halber, Thomas Funkhouser, and Matthias Niessner.
\newblock Scannet: Richly-annotated 3d reconstructions of indoor scenes, 2017.

\bibitem[Dasari et~al.(2019)Dasari, Ebert, Tian, Nair, Bucher, Schmeckpeper, Singh, Levine, and Finn]{dasari2019robonet}
Sudeep Dasari, Frederik Ebert, Stephen Tian, Suraj Nair, Bernadette Bucher, Karl Schmeckpeper, Siddharth Singh, Sergey Levine, and Chelsea Finn.
\newblock Robonet: Large-scale multi-robot learning.
\newblock In \emph{CoRL 2019: Volume 100 Proceedings of Machine Learning Research}, 2019.

\bibitem[Debevec et~al.(2000)Debevec, Hawkins, Tchou, Duiker, Sarokin, and Sagar]{debevec2000acquiring}
Paul Debevec, Tim Hawkins, Chris Tchou, Haarm-Pieter Duiker, Westley Sarokin, and Mark Sagar.
\newblock Acquiring the reflectance field of a human face, 2000.

\bibitem[Downs et~al.(2022)Downs, Francis, Koenig, Kinman, Hickman, Reymann, McHugh, and Vanhoucke]{downs2022google}
Laura Downs, Anthony Francis, Nate Koenig, Brandon Kinman, Ryan Hickman, Krista Reymann, Thomas~B. McHugh, and Vincent Vanhoucke.
\newblock Google scanned objects: A high-quality dataset of 3d scanned household items.
\newblock 2022.

\bibitem[Driess et~al.(2022)Driess, Schubert, Florence, Li, and Toussaint]{driess2022reinforcement}
Danny Driess, Ingmar Schubert, Pete Florence, Yunzhu Li, and Marc Toussaint.
\newblock Reinforcement learning with neural radiance fields.
\newblock \emph{Advances in Neural Information Processing Systems}, 35:\penalty0 16931--16945, 2022.

\bibitem[Driess et~al.(2023)Driess, Huang, Li, Tedrake, and Toussaint]{driess2023learning}
Danny Driess, Zhiao Huang, Yunzhu Li, Russ Tedrake, and Marc Toussaint.
\newblock Learning multi-object dynamics with compositional neural radiance fields.
\newblock In \emph{Conference on Robot Learning}, pages 1755--1768. PMLR, 2023.

\bibitem[Du et~al.(2021)Du, Zhang, Yu, Tenenbaum, and Wu]{du2021neural}
Yilun Du, Yinan Zhang, Hong-Xing Yu, Joshua~B Tenenbaum, and Jiajun Wu.
\newblock Neural radiance flow for 4d view synthesis and video processing.
\newblock In \emph{2021 IEEE/CVF International Conference on Computer Vision (ICCV)}, pages 14304--14314. IEEE Computer Society, 2021.

\bibitem[Fang et~al.(2023)Fang, Fang, Tang, Liu, Wang, Zhu, and Lu]{fang2023rh20t}
Hao-Shu Fang, Hongjie Fang, Zhenyu Tang, Jirong Liu, Junbo Wang, Haoyi Zhu, and Cewu Lu.
\newblock Rh20t: A robotic dataset for learning diverse skills in one-shot.
\newblock \emph{arXiv preprint arXiv:2307.00595}, 2023.

\bibitem[Fang et~al.(2022)Fang, Yi, Wang, Xie, Zhang, Liu, Nie{\ss}ner, and Tian]{fang2022fast}
Jiemin Fang, Taoran Yi, Xinggang Wang, Lingxi Xie, Xiaopeng Zhang, Wenyu Liu, Matthias Nie{\ss}ner, and Qi Tian.
\newblock Fast dynamic radiance fields with time-aware neural voxels.
\newblock In \emph{SIGGRAPH Asia 2022 Conference Papers}, pages 1--9, 2022.

\bibitem[Fridovich-Keil et~al.(2023)Fridovich-Keil, Meanti, Warburg, Recht, and Kanazawa]{fridovich2023k}
Sara Fridovich-Keil, Giacomo Meanti, Frederik~Rahb{\ae}k Warburg, Benjamin Recht, and Angjoo Kanazawa.
\newblock K-planes: Explicit radiance fields in space, time, and appearance.
\newblock In \emph{Proceedings of the IEEE/CVF Conference on Computer Vision and Pattern Recognition}, pages 12479--12488, 2023.

\bibitem[Gao et~al.(2021)Gao, Saraf, Kopf, and Huang]{gao2021dynamic}
Chen Gao, Ayush Saraf, Johannes Kopf, and Jia-Bin Huang.
\newblock Dynamic view synthesis from dynamic monocular video.
\newblock In \emph{Proceedings of the IEEE/CVF International Conference on Computer Vision}, pages 5712--5721, 2021.

\bibitem[Garrido-Jurado et~al.(2014)Garrido-Jurado, Mu{\~n}oz-Salinas, Madrid-Cuevas, and Mar{\'\i}n-Jim{\'e}nez]{garrido2014automatic}
Sergio Garrido-Jurado, Rafael Mu{\~n}oz-Salinas, Francisco~Jos{\'e} Madrid-Cuevas, and Manuel~Jes{\'u}s Mar{\'\i}n-Jim{\'e}nez.
\newblock Automatic generation and detection of highly reliable fiducial markers under occlusion.
\newblock \emph{Pattern Recognition}, 47\penalty0 (6):\penalty0 2280--2292, 2014.

\bibitem[Gu et~al.(2023)Gu, Xiang, Li, Ling, Liu, Mu, Tang, Tao, Wei, Yao, Yuan, Xie, Huang, Chen, and Su]{gu2023maniskill2}
Jiayuan Gu, Fanbo Xiang, Xuanlin Li, Zhan Ling, Xiqiang Liu, Tongzhou Mu, Yihe Tang, Stone Tao, Xinyue Wei, Yunchao Yao, Xiaodi Yuan, Pengwei Xie, Zhiao Huang, Rui Chen, and Hao Su.
\newblock Maniskill2: A unified benchmark for generalizable manipulation skills.
\newblock In \emph{International Conference on Learning Representations}, 2023.

\bibitem[Ha and Schmidhuber(2018)]{ha2018world}
David Ha and J{\"u}rgen Schmidhuber.
\newblock World models.
\newblock \emph{arXiv preprint arXiv:1803.10122}, 2018.

\bibitem[Huang et~al.(2021)Huang, Hu, Du, Zhou, Su, Tenenbaum, and Gan]{huang2021plasticinelab}
Zhiao Huang, Yuanming Hu, Tao Du, Siyuan Zhou, Hao Su, Joshua~B Tenenbaum, and Chuang Gan.
\newblock Plasticinelab: A soft-body manipulation benchmark with differentiable physics.
\newblock \emph{arXiv preprint arXiv:2104.03311}, 2021.

\bibitem[James et~al.(2020)James, Ma, Arrojo, and Davison]{james2020rlbench}
Stephen James, Zicong Ma, David~Rovick Arrojo, and Andrew~J Davison.
\newblock Rlbench: The robot learning benchmark \& learning environment.
\newblock \emph{IEEE Robotics and Automation Letters}, 5\penalty0 (2):\penalty0 3019--3026, 2020.

\bibitem[Jensen et~al.(2014)Jensen, Dahl, Vogiatzis, Tola, and Aan{\ae}s]{jensen2014large}
Rasmus Jensen, Anders Dahl, George Vogiatzis, Engin Tola, and Henrik Aan{\ae}s.
\newblock Large scale multi-view stereopsis evaluation.
\newblock In \emph{Proceedings of the IEEE conference on computer vision and pattern recognition}, pages 406--413, 2014.

\bibitem[Joo et~al.(2015)Joo, Liu, Tan, Gui, Nabbe, Matthews, Kanade, Nobuhara, and Sheikh]{joo2015panoptic}
Hanbyul Joo, Hao Liu, Lei Tan, Lin Gui, Bart Nabbe, Iain Matthews, Takeo Kanade, Shohei Nobuhara, and Yaser Sheikh.
\newblock Panoptic studio: A massively multiview system for social motion capture.
\newblock In \emph{Proceedings of the IEEE International Conference on Computer Vision}, pages 3334--3342, 2015.

\bibitem[Kerbl et~al.(2023)Kerbl, Kopanas, Leimk{\"u}hler, and Drettakis]{kerbl20233d}
Bernhard Kerbl, Georgios Kopanas, Thomas Leimk{\"u}hler, and George Drettakis.
\newblock 3d gaussian splatting for real-time radiance field rendering.
\newblock \emph{ACM Transactions on Graphics (ToG)}, 42\penalty0 (4):\penalty0 1--14, 2023.

\bibitem[Li et~al.(2023{\natexlab{a}})Li, Zhang, Wong, Gokmen, Srivastava, Mart{\'\i}n-Mart{\'\i}n, Wang, Levine, Lingelbach, Sun, et~al.]{li2023behavior}
Chengshu Li, Ruohan Zhang, Josiah Wong, Cem Gokmen, Sanjana Srivastava, Roberto Mart{\'\i}n-Mart{\'\i}n, Chen Wang, Gabrael Levine, Michael Lingelbach, Jiankai Sun, et~al.
\newblock Behavior-1k: A benchmark for embodied ai with 1,000 everyday activities and realistic simulation.
\newblock In \emph{Conference on Robot Learning}, pages 80--93. PMLR, 2023{\natexlab{a}}.

\bibitem[Li et~al.(2023{\natexlab{b}})Li, Qiao, Chen, Jatavallabhula, Lin, Jiang, and Gan]{li2023pac}
Xuan Li, Yi-Ling Qiao, Peter~Yichen Chen, Krishna~Murthy Jatavallabhula, Ming Lin, Chenfanfu Jiang, and Chuang Gan.
\newblock Pac-nerf: Physics augmented continuum neural radiance fields for geometry-agnostic system identification.
\newblock \emph{arXiv preprint arXiv:2303.05512}, 2023{\natexlab{b}}.

\bibitem[Li et~al.(2018)Li, Wu, Tedrake, Tenenbaum, and Torralba]{li2018learning}
Yunzhu Li, Jiajun Wu, Russ Tedrake, Joshua~B Tenenbaum, and Antonio Torralba.
\newblock Learning particle dynamics for manipulating rigid bodies, deformable objects, and fluids.
\newblock \emph{arXiv preprint arXiv:1810.01566}, 2018.

\bibitem[Li et~al.(2022)Li, Li, Sitzmann, Agrawal, and Torralba]{pmlr-v164-li22a}
Yunzhu Li, Shuang Li, Vincent Sitzmann, Pulkit Agrawal, and Antonio Torralba.
\newblock 3d neural scene representations for visuomotor control.
\newblock In \emph{Proceedings of the 5th Conference on Robot Learning}, pages 112--123. PMLR, 2022.

\bibitem[Li et~al.(2021)Li, Niklaus, Snavely, and Wang]{li2021neural}
Zhengqi Li, Simon Niklaus, Noah Snavely, and Oliver Wang.
\newblock Neural scene flow fields for space-time view synthesis of dynamic scenes.
\newblock In \emph{Proceedings of the IEEE/CVF Conference on Computer Vision and Pattern Recognition}, pages 6498--6508, 2021.

\bibitem[Lin et~al.(2021{\natexlab{a}})Lin, Xiao, Liu, Yang, and Ramamoorthi]{lin2021deep}
Kai-En Lin, Lei Xiao, Feng Liu, Guowei Yang, and Ravi Ramamoorthi.
\newblock Deep 3d mask volume for view synthesis of dynamic scenes.
\newblock In \emph{ICCV}, 2021{\natexlab{a}}.

\bibitem[Lin et~al.(2021{\natexlab{b}})Lin, Wang, Olkin, and Held]{lin2021softgym}
Xingyu Lin, Yufei Wang, Jake Olkin, and David Held.
\newblock Softgym: Benchmarking deep reinforcement learning for deformable object manipulation.
\newblock In \emph{Conference on Robot Learning}, pages 432--448. PMLR, 2021{\natexlab{b}}.

\bibitem[Ling et~al.(2023)Ling, Yao, Li, and Su]{ling2023efficacy}
Zhan Ling, Yunchao Yao, Xuanlin Li, and Hao Su.
\newblock On the efficacy of 3d point cloud reinforcement learning.
\newblock \emph{arXiv preprint arXiv:2306.06799}, 2023.

\bibitem[Liu et~al.(2023)Liu, Chen, Fu, Wu, Jin, Li, Wong, Xu, Ramamoorthi, Xu, and Su]{liu2023openillumination}
Isabella Liu, Linghao Chen, Ziyang Fu, Liwen Wu, Haian Jin, Zhong Li, Chin Ming~Ryan Wong, Yi Xu, Ravi Ramamoorthi, Zexiang Xu, and Hao Su.
\newblock Openillumination: A multi-illumination dataset for inverse rendering evaluation on real objects.
\newblock 2023.

\bibitem[Liu et~al.(2022)Liu, Liu, Jiang, Lyu, Wan, Shen, Liang, Fu, Wang, and Yi]{liu2022hoi4d}
Yunze Liu, Yun Liu, Che Jiang, Kangbo Lyu, Weikang Wan, Hao Shen, Boqiang Liang, Zhoujie Fu, He Wang, and Li Yi.
\newblock Hoi4d: A 4d egocentric dataset for category-level human-object interaction.
\newblock In \emph{Proceedings of the IEEE/CVF Conference on Computer Vision and Pattern Recognition}, pages 21013--21022, 2022.

\bibitem[Lombardi et~al.(2019)Lombardi, Simon, Saragih, Schwartz, Lehrmann, and Sheikh]{lombardi2019neural}
Stephen Lombardi, Tomas Simon, Jason Saragih, Gabriel Schwartz, Andreas Lehrmann, and Yaser Sheikh.
\newblock Neural volumes: Learning dynamic renderable volumes from images.
\newblock \emph{ACM Transactions on Graphics (SIGGRAPH 2019) 38, 4, Article 65}, 38\penalty0 (4):\penalty0 1--14, 2019.

\bibitem[Lu et~al.(2023)Lu, Zhou, Xing, Pokhariya, Dey, Shah, Mavidipalli, Hu, Comport, Chen, et~al.]{lu2023diva}
Cheng-You Lu, Peisen Zhou, Angela Xing, Chandradeep Pokhariya, Arnab Dey, Ishaan Shah, Rugved Mavidipalli, Dylan Hu, Andrew Comport, Kefan Chen, et~al.
\newblock Diva-360: The dynamic visuo-audio dataset for immersive neural fields.
\newblock \emph{arXiv preprint arXiv:2307.16897}, 2023.

\bibitem[Luiten et~al.(2023)Luiten, Kopanas, Leibe, and Ramanan]{luiten2023dynamic}
Jonathon Luiten, Georgios Kopanas, Bastian Leibe, and Deva Ramanan.
\newblock Dynamic 3d gaussians: Tracking by persistent dynamic view synthesis.
\newblock \emph{arXiv preprint arXiv:2308.09713}, 2023.

\bibitem[Mandi et~al.(2022)Mandi, Liu, Lee, and Abbeel]{mandi2022towards}
Zhao Mandi, Fangchen Liu, Kimin Lee, and Pieter Abbeel.
\newblock Towards more generalizable one-shot visual imitation learning.
\newblock In \emph{2022 International Conference on Robotics and Automation (ICRA)}, pages 2434--2444. IEEE, 2022.

\bibitem[Mandlekar et~al.(2018)Mandlekar, Zhu, Garg, Booher, Spero, Tung, Gao, Emmons, Gupta, Orbay, et~al.]{mandlekar2018roboturk}
Ajay Mandlekar, Yuke Zhu, Animesh Garg, Jonathan Booher, Max Spero, Albert Tung, Julian Gao, John Emmons, Anchit Gupta, Emre Orbay, et~al.
\newblock Roboturk: A crowdsourcing platform for robotic skill learning through imitation.
\newblock In \emph{Conference on Robot Learning}, pages 879--893. PMLR, 2018.

\bibitem[Mandlekar et~al.(2021)Mandlekar, Xu, Wong, Nasiriany, Wang, Kulkarni, Fei-Fei, Savarese, Zhu, and Mart{\'\i}n-Mart{\'\i}n]{mandlekar2021matters}
Ajay Mandlekar, Danfei Xu, Josiah Wong, Soroush Nasiriany, Chen Wang, Rohun Kulkarni, Li Fei-Fei, Silvio Savarese, Yuke Zhu, and Roberto Mart{\'\i}n-Mart{\'\i}n.
\newblock What matters in learning from offline human demonstrations for robot manipulation.
\newblock \emph{arXiv preprint arXiv:2108.03298}, 2021.

\bibitem[Mees et~al.(2022)Mees, Hermann, Rosete-Beas, and Burgard]{mees2022calvin}
Oier Mees, Lukas Hermann, Erick Rosete-Beas, and Wolfram Burgard.
\newblock Calvin: A benchmark for language-conditioned policy learning for long-horizon robot manipulation tasks.
\newblock \emph{IEEE Robotics and Automation Letters}, 7\penalty0 (3):\penalty0 7327--7334, 2022.

\bibitem[Mildenhall et~al.(2021)Mildenhall, Srinivasan, Tancik, Barron, Ramamoorthi, and Ng]{mildenhall2021nerf}
Ben Mildenhall, Pratul~P Srinivasan, Matthew Tancik, Jonathan~T Barron, Ravi Ramamoorthi, and Ren Ng.
\newblock Nerf: Representing scenes as neural radiance fields for view synthesis.
\newblock \emph{Communications of the ACM}, 65\penalty0 (1):\penalty0 99--106, 2021.

\bibitem[Park et~al.(2021{\natexlab{a}})Park, Sinha, Barron, Bouaziz, Goldman, Seitz, and Martin-Brualla]{park2021nerfies}
Keunhong Park, Utkarsh Sinha, Jonathan~T Barron, Sofien Bouaziz, Dan~B Goldman, Steven~M Seitz, and Ricardo Martin-Brualla.
\newblock Nerfies: Deformable neural radiance fields.
\newblock In \emph{Proceedings of the IEEE/CVF International Conference on Computer Vision}, pages 5865--5874, 2021{\natexlab{a}}.

\bibitem[Park et~al.(2021{\natexlab{b}})Park, Sinha, Hedman, Barron, Bouaziz, Goldman, Martin-Brualla, and Seitz]{park2021hypernerf}
Keunhong Park, Utkarsh Sinha, Peter Hedman, Jonathan~T Barron, Sofien Bouaziz, Dan~B Goldman, Ricardo Martin-Brualla, and Steven~M Seitz.
\newblock Hypernerf: A higher-dimensional representation for topologically varying neural radiance fields.
\newblock \emph{arXiv preprint arXiv:2106.13228}, 2021{\natexlab{b}}.

\bibitem[Peng et~al.(2021)Peng, Zhang, Xu, Wang, Shuai, Bao, and Zhou]{peng2021neural}
Sida Peng, Yuanqing Zhang, Yinghao Xu, Qianqian Wang, Qing Shuai, Hujun Bao, and Xiaowei Zhou.
\newblock Neural body: Implicit neural representations with structured latent codes for novel view synthesis of dynamic humans.
\newblock In \emph{Proceedings of the IEEE/CVF Conference on Computer Vision and Pattern Recognition}, pages 9054--9063, 2021.

\bibitem[Pumarola et~al.(2021)Pumarola, Corona, Pons-Moll, and Moreno-Noguer]{pumarola2021d}
Albert Pumarola, Enric Corona, Gerard Pons-Moll, and Francesc Moreno-Noguer.
\newblock D-nerf: Neural radiance fields for dynamic scenes.
\newblock In \emph{Proceedings of the IEEE/CVF Conference on Computer Vision and Pattern Recognition}, pages 10318--10327, 2021.

\bibitem[Qin et~al.(2023)Qin, Yang, Huang, Van~Wyk, Su, Wang, Chao, and Fox]{qin2023anyteleop}
Yuzhe Qin, Wei Yang, Binghao Huang, Karl Van~Wyk, Hao Su, Xiaolong Wang, Yu-Wei Chao, and Dietor Fox.
\newblock Anyteleop: A general vision-based dexterous robot arm-hand teleoperation system.
\newblock \emph{arXiv preprint arXiv:2307.04577}, 2023.

\bibitem[Sanchez-Gonzalez et~al.(2020)Sanchez-Gonzalez, Godwin, Pfaff, Ying, Leskovec, and Battaglia]{sanchez2020learning}
Alvaro Sanchez-Gonzalez, Jonathan Godwin, Tobias Pfaff, Rex Ying, Jure Leskovec, and Peter Battaglia.
\newblock Learning to simulate complex physics with graph networks.
\newblock In \emph{International conference on machine learning}, pages 8459--8468. PMLR, 2020.

\bibitem[Sch{\"o}nberger et~al.(2016)Sch{\"o}nberger, Zheng, Frahm, and Pollefeys]{schonberger2016pixelwise}
Johannes~L Sch{\"o}nberger, Enliang Zheng, Jan-Michael Frahm, and Marc Pollefeys.
\newblock Pixelwise view selection for unstructured multi-view stereo.
\newblock In \emph{Computer Vision--ECCV 2016: 14th European Conference, Amsterdam, The Netherlands, October 11-14, 2016, Proceedings, Part III 14}, pages 501--518. Springer, 2016.

\bibitem[Seo et~al.(2023)Seo, Kim, James, Lee, Shin, and Abbeel]{seo2023multi}
Younggyo Seo, Junsu Kim, Stephen James, Kimin Lee, Jinwoo Shin, and Pieter Abbeel.
\newblock Multi-view masked world models for visual robotic manipulation.
\newblock \emph{arXiv preprint arXiv:2302.02408}, 2023.

\bibitem[Shao et~al.(2023)Shao, Zheng, Tu, Liu, Zhang, and Liu]{shao2023tensor4d}
Ruizhi Shao, Zerong Zheng, Hanzhang Tu, Boning Liu, Hongwen Zhang, and Yebin Liu.
\newblock Tensor4d: Efficient neural 4d decomposition for high-fidelity dynamic reconstruction and rendering.
\newblock In \emph{Proceedings of the IEEE/CVF Conference on Computer Vision and Pattern Recognition}, pages 16632--16642, 2023.

\bibitem[Sharma et~al.(2018)Sharma, Mohan, Pinto, and Gupta]{sharma2018multiple}
Pratyusha Sharma, Lekha Mohan, Lerrel Pinto, and Abhinav Gupta.
\newblock Multiple interactions made easy (mime): Large scale demonstrations data for imitation.
\newblock In \emph{Conference on robot learning}, pages 906--915. PMLR, 2018.

\bibitem[Shim et~al.(2023)Shim, Lee, and Kim]{shim2023snerl}
Dongseok Shim, Seungjae Lee, and H~Jin Kim.
\newblock Snerl: Semantic-aware neural radiance fields for reinforcement learning.
\newblock \emph{arXiv preprint arXiv:2301.11520}, 2023.

\bibitem[Silberman et~al.(2012)Silberman, Hoiem, Kohli, and Fergus]{silberman2012indoor}
Nathan Silberman, Derek Hoiem, Pushmeet Kohli, and Rob Fergus.
\newblock Indoor segmentation and support inference from rgbd images, 2012.

\bibitem[Singh et~al.(2014)Singh, Sha, Narayan, Achim, and Abbeel]{singh2014bigbird}
Arjun Singh, James Sha, Karthik~S. Narayan, Tudor Achim, and Pieter Abbeel.
\newblock Bigbird: A large-scale 3d database of object instances, 2014.

\bibitem[Song et~al.(2015)Song, Lichtenberg, and Xiao]{song2015sun}
Shuran Song, Samuel~P. Lichtenberg, and Jianxiong Xiao.
\newblock Sun rgb-d: A rgb-d scene understanding benchmark suite, 2015.

\bibitem[Torabi et~al.(2018)Torabi, Warnell, and Stone]{torabi2018behavioral}
Faraz Torabi, Garrett Warnell, and Peter Stone.
\newblock Behavioral cloning from observation.
\newblock \emph{arXiv preprint arXiv:1805.01954}, 2018.

\bibitem[Toschi et~al.(2023)Toschi, De~Matteo, Spezialetti, De~Gregorio, Di~Stefano, and Salti]{toschi2023relight}
Marco Toschi, Riccardo De~Matteo, Riccardo Spezialetti, Daniele De~Gregorio, Luigi Di~Stefano, and Samuele Salti.
\newblock Relight my nerf: A dataset for novel view synthesis and relighting of real world objects.
\newblock In \emph{Proceedings of the IEEE/CVF Conference on Computer Vision and Pattern Recognition}, pages 20762--20772, 2023.

\bibitem[Tretschk et~al.(2021)Tretschk, Tewari, Golyanik, Zollh{\"o}fer, Lassner, and Theobalt]{tretschk2021non}
Edgar Tretschk, Ayush Tewari, Vladislav Golyanik, Michael Zollh{\"o}fer, Christoph Lassner, and Christian Theobalt.
\newblock Non-rigid neural radiance fields: Reconstruction and novel view synthesis of a dynamic scene from monocular video.
\newblock In \emph{Proceedings of the IEEE/CVF International Conference on Computer Vision}, pages 12959--12970, 2021.

\bibitem[Vaswani et~al.(2017)Vaswani, Shazeer, Parmar, Uszkoreit, Jones, Gomez, Kaiser, and Polosukhin]{vaswani2017attention}
Ashish Vaswani, Noam Shazeer, Niki Parmar, Jakob Uszkoreit, Llion Jones, Aidan~N Gomez, {\L}ukasz Kaiser, and Illia Polosukhin.
\newblock Attention is all you need.
\newblock \emph{Advances in neural information processing systems}, 30, 2017.

\bibitem[Walke et~al.(2023)Walke, Black, Lee, Kim, Du, Zheng, Zhao, Hansen-Estruch, Vuong, He, Myers, Fang, Finn, and Levine]{walke2023bridgedata}
Homer Walke, Kevin Black, Abraham Lee, Moo~Jin Kim, Max Du, Chongyi Zheng, Tony Zhao, Philippe Hansen-Estruch, Quan Vuong, Andre He, Vivek Myers, Kuan Fang, Chelsea Finn, and Sergey Levine.
\newblock Bridgedata v2: A dataset for robot learning at scale.
\newblock In \emph{Conference on Robot Learning (CoRL)}, 2023.

\bibitem[Wang et~al.(2023)Wang, Li, Zhang, Driggs-Campbell, Wu, Fei-Fei, and Li]{wang2023d}
Yixuan Wang, Zhuoran Li, Mingtong Zhang, Katherine Driggs-Campbell, Jiajun Wu, Li Fei-Fei, and Yunzhu Li.
\newblock D3fields: Dynamic 3d descriptor fields for zero-shot generalizable robotic manipulation.
\newblock \emph{arXiv preprint arXiv:2309.16118}, 2023.

\bibitem[Wu et~al.(2023)Wu, Yi, Fang, Xie, Zhang, Wei, Liu, Tian, and Wang]{wu20234d}
Guanjun Wu, Taoran Yi, Jiemin Fang, Lingxi Xie, Xiaopeng Zhang, Wei Wei, Wenyu Liu, Qi Tian, and Xinggang Wang.
\newblock 4d gaussian splatting for real-time dynamic scene rendering.
\newblock \emph{arXiv preprint arXiv:2310.08528}, 2023.

\bibitem[Xian et~al.(2021)Xian, Huang, Kopf, and Kim]{xian2021space}
Wenqi Xian, Jia-Bin Huang, Johannes Kopf, and Changil Kim.
\newblock Space-time neural irradiance fields for free-viewpoint video.
\newblock In \emph{Proceedings of the IEEE/CVF Conference on Computer Vision and Pattern Recognition}, pages 9421--9431, 2021.

\bibitem[Xian et~al.(2023)Xian, Zhu, Xu, Tung, Torralba, Fragkiadaki, and Gan]{xian2023fluidlab}
Zhou Xian, Bo Zhu, Zhenjia Xu, Hsiao-Yu Tung, Antonio Torralba, Katerina Fragkiadaki, and Chuang Gan.
\newblock Fluidlab: A differentiable environment for benchmarking complex fluid manipulation.
\newblock \emph{arXiv preprint arXiv:2303.02346}, 2023.

\bibitem[Yang et~al.(2023)Yang, Gao, Zhou, Jiao, Zhang, and Jin]{yang2023deformable}
Ziyi Yang, Xinyu Gao, Wen Zhou, Shaohui Jiao, Yuqing Zhang, and Xiaogang Jin.
\newblock Deformable 3d gaussians for high-fidelity monocular dynamic scene reconstruction.
\newblock \emph{arXiv preprint arXiv:2309.13101}, 2023.

\bibitem[Yu et~al.(2020)Yu, Quillen, He, Julian, Hausman, Finn, and Levine]{yu2020meta}
Tianhe Yu, Deirdre Quillen, Zhanpeng He, Ryan Julian, Karol Hausman, Chelsea Finn, and Sergey Levine.
\newblock Meta-world: A benchmark and evaluation for multi-task and meta reinforcement learning.
\newblock In \emph{Conference on robot learning}, pages 1094--1100. PMLR, 2020.

\bibitem[Ze et~al.(2023{\natexlab{a}})Ze, Hansen, Chen, Jain, and Wang]{ze2023visual}
Yanjie Ze, Nicklas Hansen, Yinbo Chen, Mohit Jain, and Xiaolong Wang.
\newblock Visual reinforcement learning with self-supervised 3d representations.
\newblock \emph{IEEE Robotics and Automation Letters}, 8\penalty0 (5):\penalty0 2890--2897, 2023{\natexlab{a}}.

\bibitem[Ze et~al.(2023{\natexlab{b}})Ze, Yan, Wu, Macaluso, Ge, Ye, Hansen, Li, and Wang]{ze2023gnfactor}
Yanjie Ze, Ge Yan, Yueh-Hua Wu, Annabella Macaluso, Yuying Ge, Jianglong Ye, Nicklas Hansen, Li~Erran Li, and Xiaolong Wang.
\newblock Gnfactor: Multi-task real robot learning with generalizable neural feature fields.
\newblock \emph{arXiv preprint arXiv:2308.16891}, 2023{\natexlab{b}}.

\end{thebibliography}
}

\clearpage
\setcounter{page}{1}
\maketitlesupplementary

\section{System Specifics}
The 86 Canon 250D cameras record video at 30 FPS, 1080p resolution, 1/200 shutter speed, F8.0 aperture, ISO 3200, and Fluorescent white balance. This configuration is selected to balance image quality with motion blur.

The 3 RealSense D435 cameras record pixel-aligned color and depth video at 30 FPS at 480p, which is the recommended resolution of D435 for depth image quality.

The 86 Canon cameras are controlled by 23 Raspberry Pi 4B single-board computers for video recording and downloading. The Raspberry Pis and the main PC are connected via LAN using an internet switch. These devices with the main PC communicate within 192.168.0.x. 

The main PC controls the xArm6 robot arm and gripper via direct ethernet connection on 192.168.1.x. The main PC reads the target end-effector pose and gripper position from the VR game controller via USB, computes the target joint position, and drives the robot arm by sending a packet to the xArm control box via ethernet using the xArm factory-provided API. This control loop runs at 30 Hz.

We use 4 Daylight Artist Studio Lamps for illumination, and 1 TYCOLIT Outdoor Led Flood Light for temporal alignment.

The desk is 152.4 cm by 76.2 cm.

\section{Temporal Alignment Error Analysis}
\begin{figure}[h]
    \centering
    \includegraphics[width=0.4\textwidth]{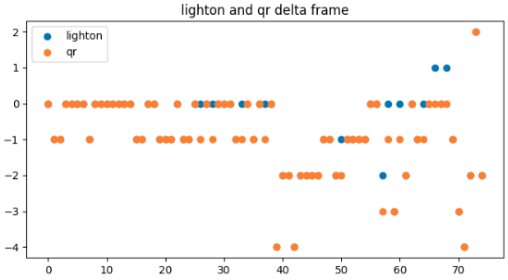}
    \caption{Temporal alignment error}
    \label{fig:temp_error}
\end{figure}
\begin{figure*}[h]
    \centering
    \includegraphics[width=1\textwidth]{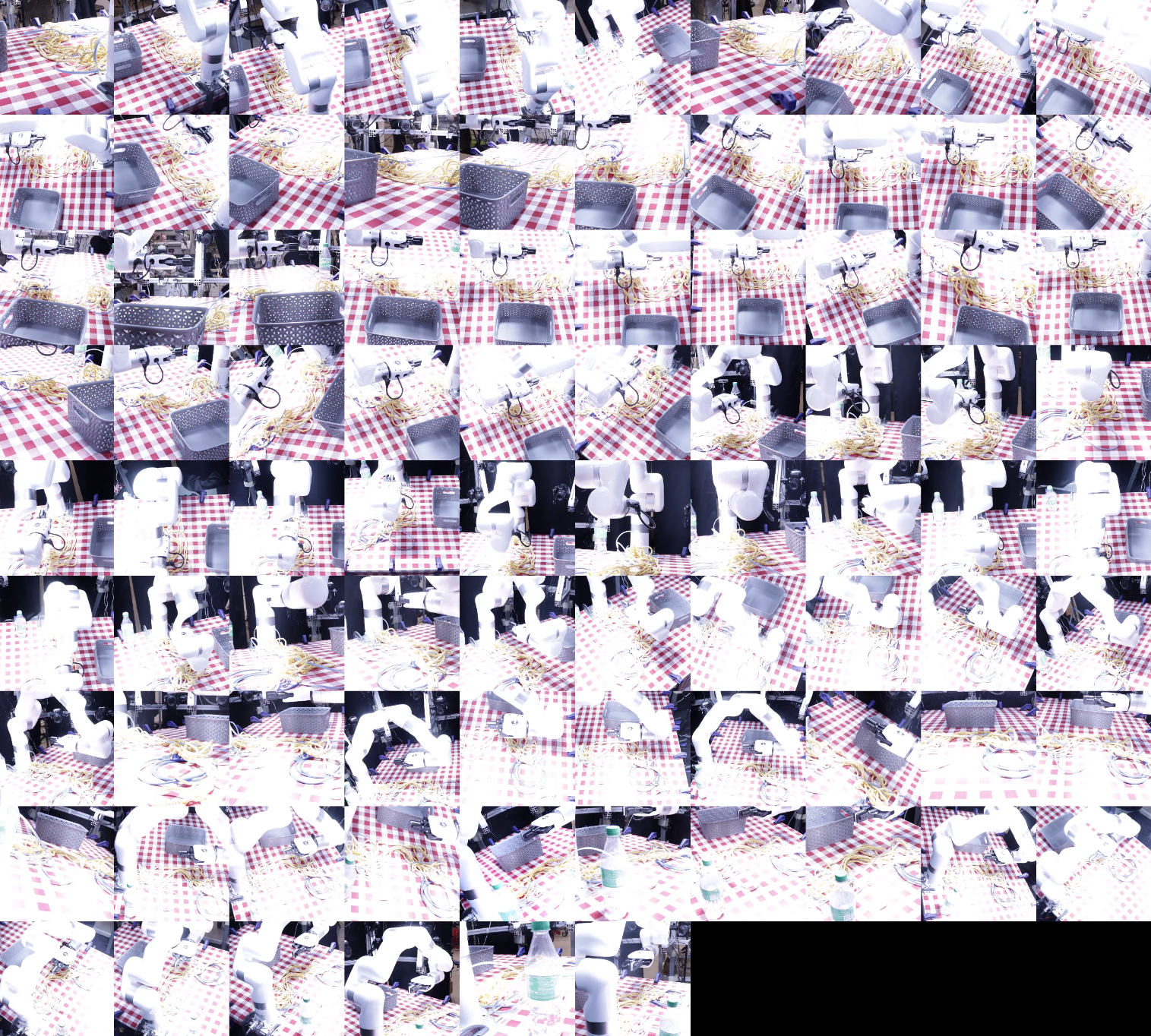}
    \caption{Example light on alignment event captured by all 86 cameras}
    \label{fig:light_on}
\end{figure*}
\begin{figure*}[h]
    \centering
    \includegraphics[width=1\textwidth]{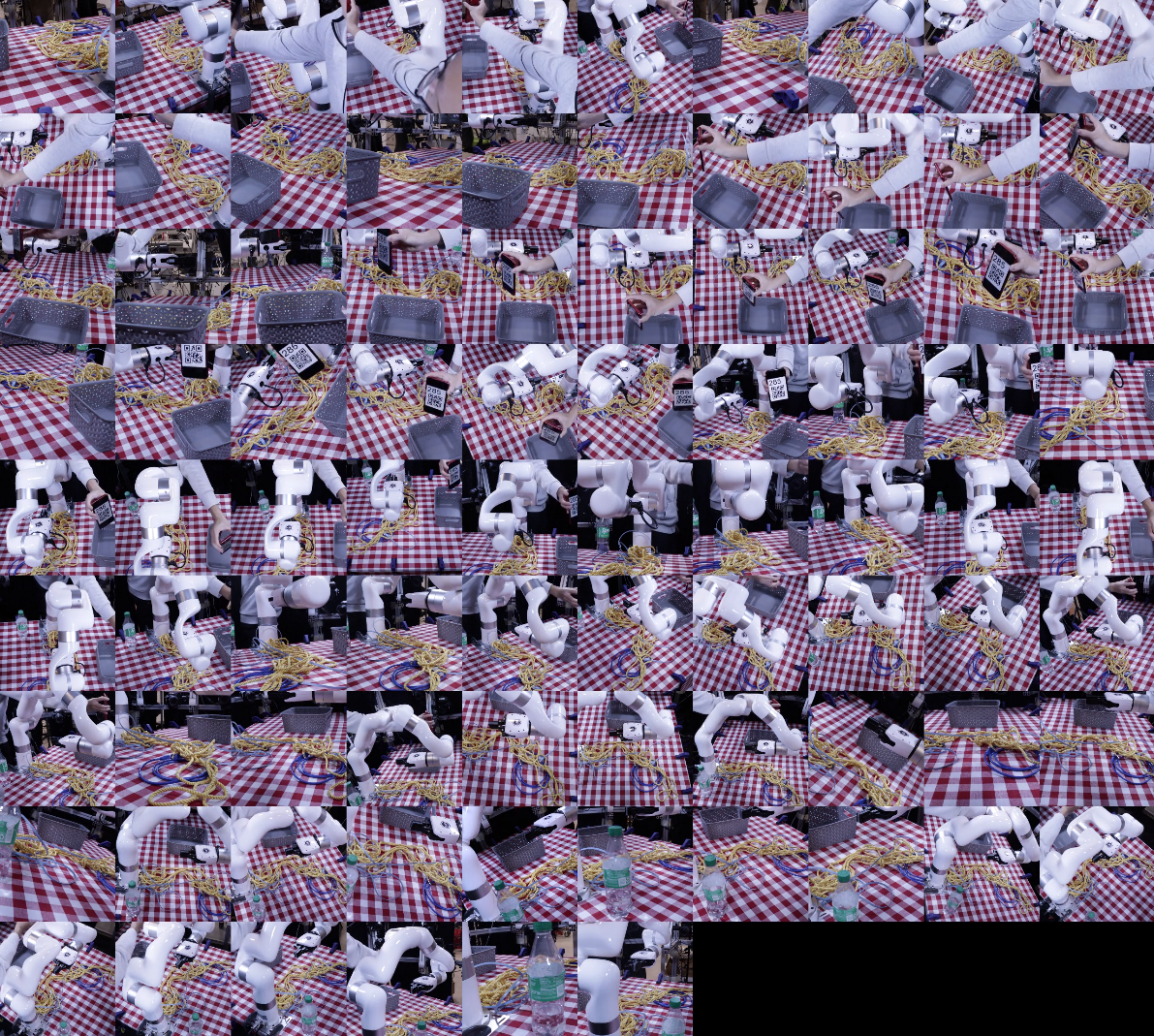}
    \caption{Example QR code alignment event captured by 86 cameras over 30 seconds. The teleoperator holds a phone playing a 30 fps QR code video around the inside of RoboStage to make sure the video recorded from all views contains a portion of the footage that records the QR code at some point during the entire 30 seconds.}
    \label
{fig:qr_code}
\end{figure*}

The sudden increase in average pixel intensity is used to identify the keyframe to align the multi-view videos. This is shown in Figure \ref{fig:light_on}. To verify the alignment quality of this method, we record a multi-view video of 1. light on and off, 2. a phone playing a 30-second 30 fps QR code video of 0-899 held around the stage to ensure all videos contain a section where the QR code is visible, shown in Figure \ref{fig:qr_code}. The starting frame is found by subtracting the QR code number from the frame index of the image. To analyze the error between the 2 methods, we recorded a video with both alignment events and found the starting frame of the videos using these 2 methods. Then, we assumed the two methods agreed with each other on the video recorded from the 0th camera. After that, we plotted the temporal shift (unit: frames) between 0th and every other camera using the 2 methods, shown in Figure \ref{fig:temp_error}, where the x-axis is the camera ID and the y-axis is the relative temporal shift (unit: frames) of this video compared to the 0th camera, estimated using flood light (blue) and QR code (orange). We found out two methods have a maximum alignment discrepancy of 1 frame in 11 out of 86 videos. We believe this error is almost negligible for any downstream applications. 

The reason QR code video is not used as the alignment event is that it is inefficient in data collection and post-processing. During data collection, we empirically found that showing the QR code video to all cameras takes 30 seconds, whereas light on and off only takes 4-5 seconds max. During post-processing, the opencv QR code reading algorithm is much slower than computing the average pixel intensity of a frame, which is simply a sum and multiply operation on an integer array.

\section{Control Frequency Requirements}
The control frequency of teleoperation is crucial to enable the teleoperator to collect more diverse tasks. In our dataset, many involve the robot arm manipulating an open container full of liquid and pouring it into another container. In a task, we tasked the teleoperator to pour a cup with water filled to 3/4 to another cup. We empirically find out the well-trained teleoperator starts to spill the liquid uncontrollably when the control frequency drops below 20. When the control frequency drops below 10, the teleoperator loses control of the cup and spills almost all the liquid in the cup. There are many tasks that require high-frequency feedback control, such as liquid container manipulation, precise insertion of objects, balancing objects in hand, etc.

\section{Dynamic NeRF experiments}

\subsection{More visualization results}

\begin{figure*}[htp]
    \includegraphics[width=0.8\textwidth]{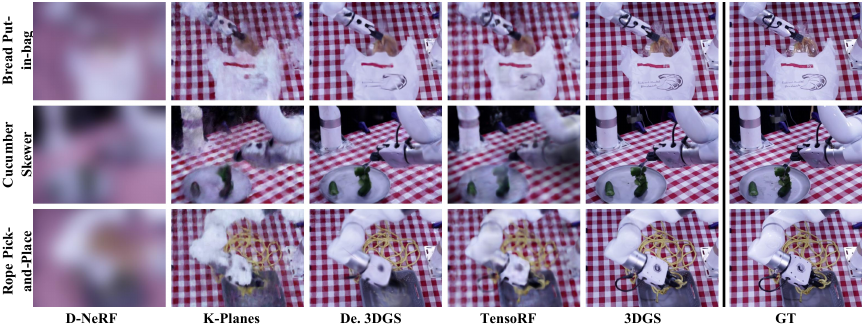}
    \centering
    \caption{Visualization of more dynamic NeRF results using 5 different baselines.}
    \label{fig:dnerf_more}
    \vspace{-0.5cm}
\end{figure*}

Here we show more qualitative results of the visualization of dynamic NeRF baselines in Figure \ref{fig:dnerf_more}.

\subsection{Video length}

To evaluate the robustness and scalability of these baseline methods, we conducted an ablation study to investigate the influence of different video lengths on the performance metrics. Specifically, we varied the length of the input video while keeping other parameters constant.  The tested video lengths included short clips (2 seconds), medium-length sequences (5 seconds), and longer videos (15 seconds). Table \ref{tab: videolength} presents the quantitative results of our ablation study for different video lengths using PSNR as the metric. The table shows that all three dynamic NeRF methods achieve the best performance on 5s-length videos, so we chose this length to evaluate the results of various methods.

\begin{center}
\begin{table}[t]
\centering
\renewcommand{\arraystretch}{1.2}
\resizebox{\linewidth}{!}{
\begin{tabular}{|c|ccc|}
\hlineB{2}
Method & 2 seconds & 5 seconds & 15 seconds  \\ 
\hline
D-NeRF     & 10.52   & 10.58 & 10.34      \\ 
K-Planes      & 16.86   & 16.99    & 15.84       \\ 
De. 3DGS & 18.42   & 18.57    & 17.81      \\ 
TensoRF       & 20.21 & 20.14 & 20.13       \\ 
3DGS     & 24.82 & 24.76 & 24.71      \\
\hlineB{2}
\end{tabular}
}
\caption{Ablation study of different video lengths: Average PSNR over all scenes.}
\label{tab: videolength}
\end{table}
\end{center}

\subsection{Training time}

We also conducted ablation studies on different training time to understand the performance of different training baselines under a limited training budget. The default training time of all 3 different dynamic NeRF methods is about 10 hours. Thus, we conducted experiments with three different training durations: 10 minutes, 1 hour, and the default 10 hours. The other parameters are kept the same, such as 5s-length video, 81 views for training and 5 views for testing, etc. The result is shown in Table \ref{tab: trainingtime}.

\begin{center}
\begin{table}[t]
\centering
\renewcommand{\arraystretch}{1.2}
\resizebox{\linewidth}{!}{
\begin{tabular}{|c|ccc|}
\hlineB{2}
Method & 10 minutes & 1 hour & 10 hours  \\ 
\hline
D-NeRF     &  8.04  & 8.83 & 10.58      \\ 
K-Planes      &  12.35  &   15.77  & 16.99       \\ 
De. 3DGS &  11.61  & 16.39 & 18.57 \\
\hlineB{2}
\end{tabular}
}
\caption{Ablation study of different training times: Average PSNR over all scenes.}
\label{tab: trainingtime}
\end{table}
\end{center}

\section{Imitation Learning}

\subsection{Task details}
\paragraph{Towel} The trained policy controls xArm6 to grasp the towel and put into the basket. The dataset contains 80 demos with each 300 steps at 30 Hz. Execution is considered a success when the towel is fully inside the basket.

\paragraph{Slippers} The trained policy controls xArm6 to move the slippers together to organize them. The dataset contains 30 demos with each 450 steps at 30 Hz. Execution is considered a success when both slippers are closer to the center than before.

\paragraph{Bottle} The trained policy controls xArm6 to flip the bottle up right. The dataset contains 200 demos with each 450 steps at 30 Hz. Execution is considered a success when the bottle is in an up right position without the robot holding it.

\paragraph{Cable} The trained policy controls xArm6 to put a blue USB extension cable into the basket. The dataset contains 150 demos with each 450 steps at 30 Hz. Execution is considered a success when the USB cable is fully inside the basket. 

\paragraph{Rope} The trained policy controls xArm6 to put a red and white rope into the basket. The dataset contains 155 demos with each 750 steps at 30 Hz. Execution is considered a success when the rope is fully inside the basket. 

\subsection{More Visualizations}
Please refer to the supplementary material video for more policy deployment visualizations.

\section{Open Source}
All data and code will be available at \url{https://robo360dataset.github.io/}.

\section{More Example Data}
\begin{figure*}[h]
    \centering
    \includegraphics[width=1\textwidth]{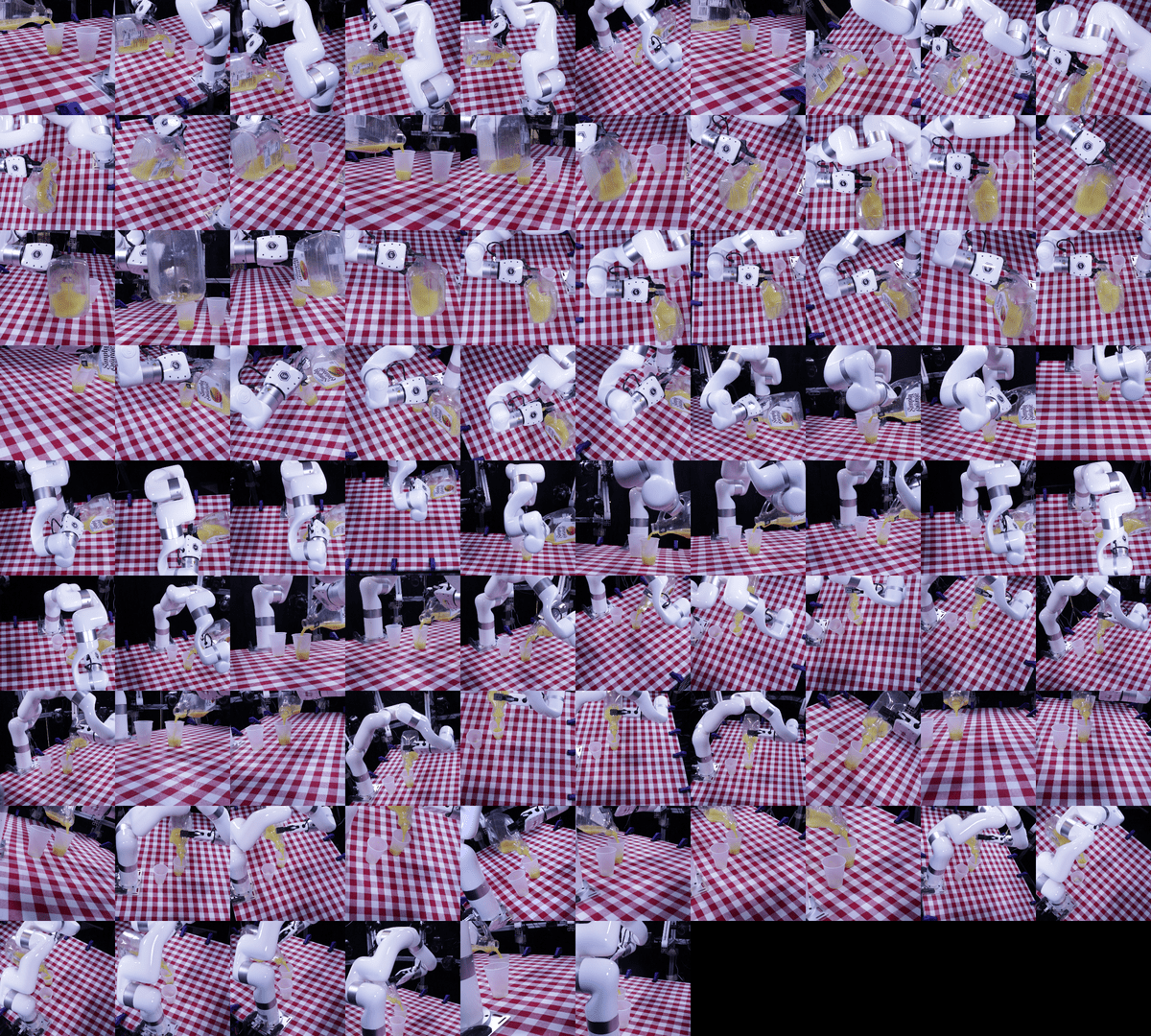}
    \caption{Video of xArm6 robot pouring orange juice into a plastic cup from all 86 views.}
    \label{fig:orange_juice}
\end{figure*}

\begin{figure*}[h]
    \centering
    \includegraphics[width=1\textwidth]{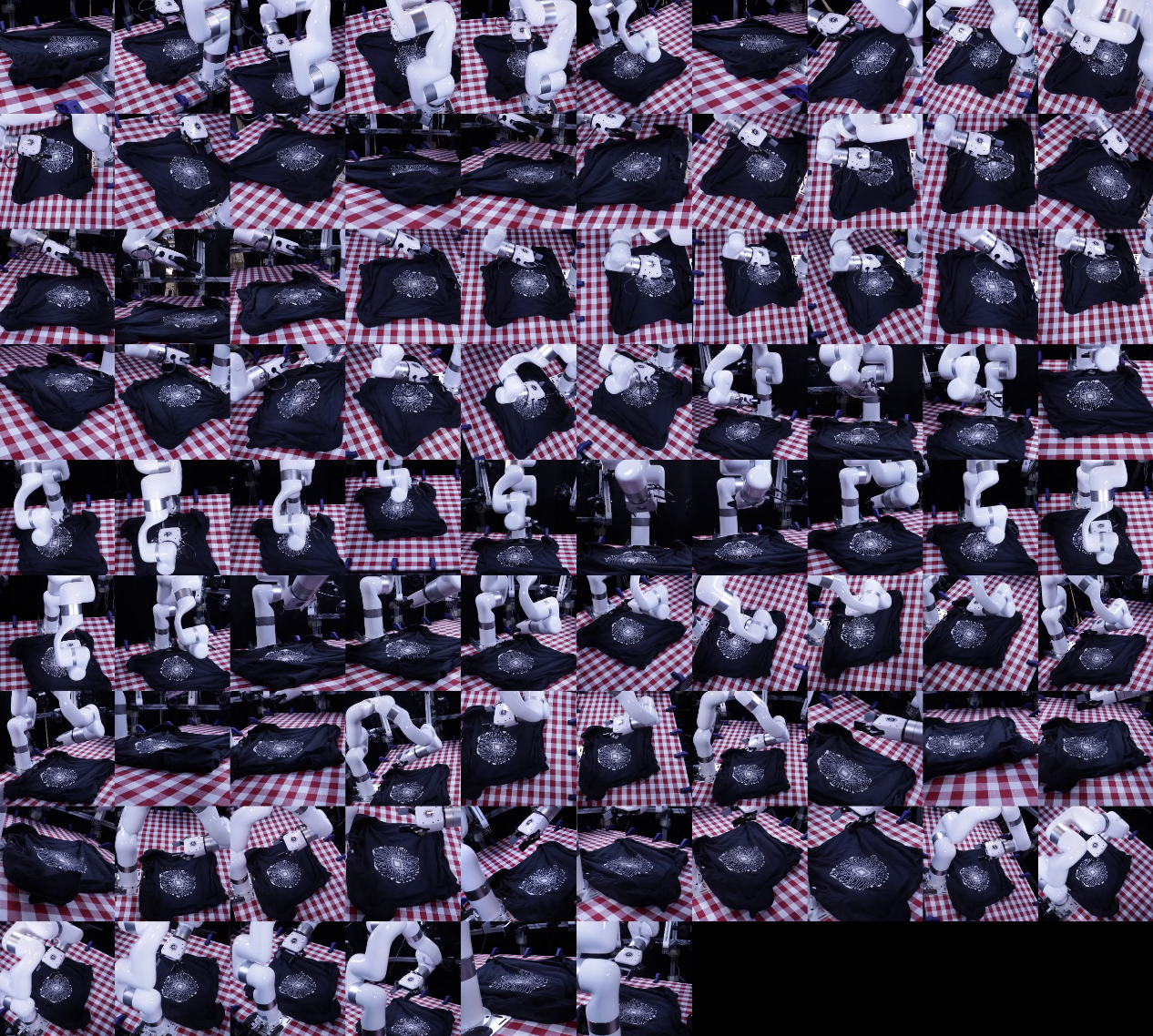}
    \caption{Video of xArm6 robot folding a t-shirt with an electric brain logo from all 86 views.}
    \label{fig:fold_tshirt}
\end{figure*}

86 views of video data of task orange juice (Figure \ref{fig:orange_juice}), fold t-shirt (Figure \ref{fig:fold_tshirt}) and sprite can tower collapse (Figure \ref{fig:sprite_tower_collapse}) are shown in the following pages. The full video can be found in the supplementary material video.
\begin{figure*}[h]
    \centering
    \includegraphics[width=1\textwidth]{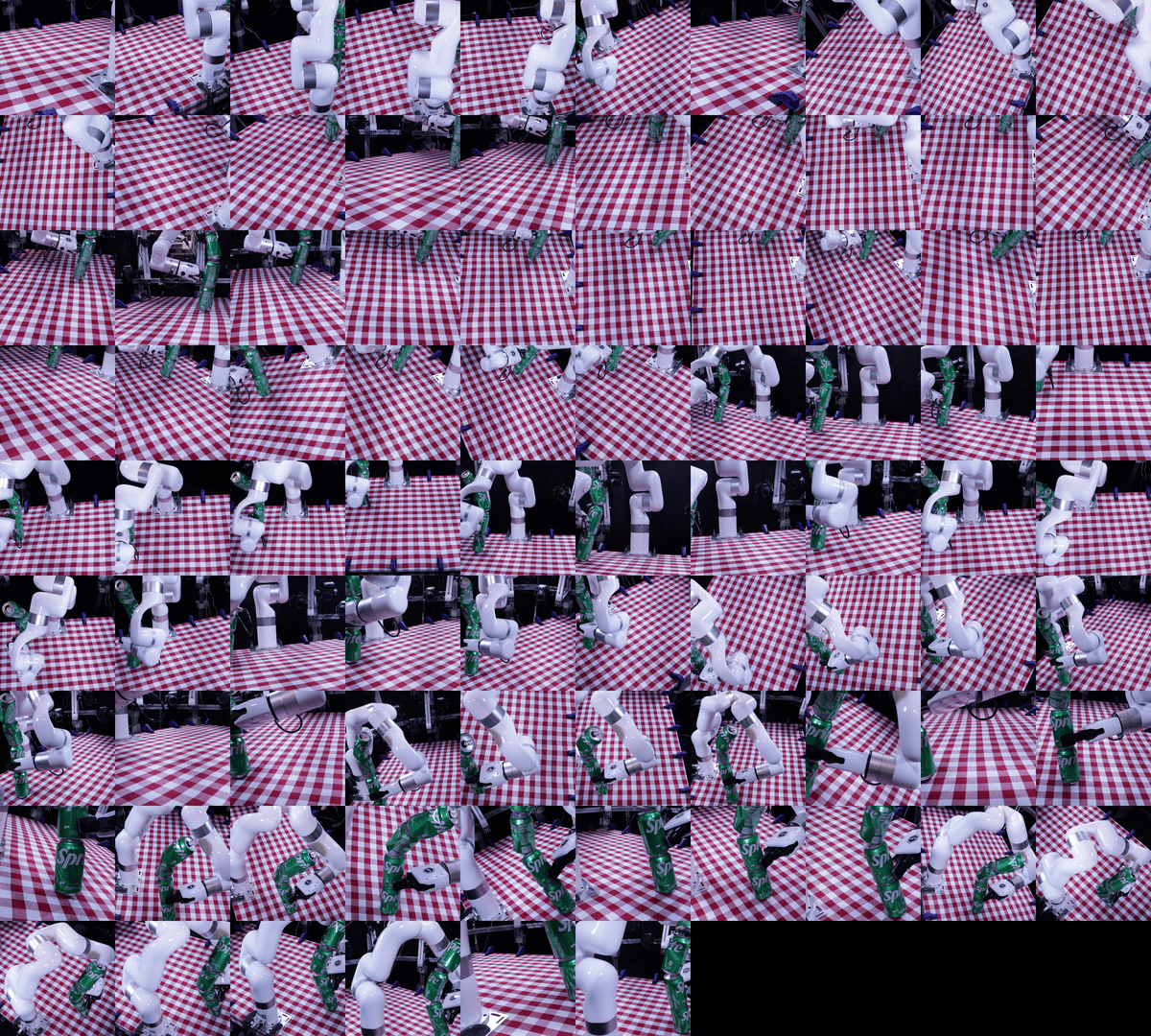}
    \caption{Video of xArm6 hitting a sprite can tower from all 86 views.}
    \label{fig:sprite_tower_collapse}
\end{figure*}

\clearpage


\end{document}